%% file: main.tex
\documentclass[10pt,twocolumn,letterpaper]{article}

\usepackage{cvpr}
\usepackage{times}
\usepackage{epsfig}
\usepackage{graphicx}
\usepackage{amsmath,amssymb,amsthm}
\usepackage{bm}
\usepackage{overpic}
\usepackage[title]{appendix}
\usepackage{pgfplots}
\usepackage{pgf}
\usepackage{tikz}

\newcommand\comment[1]{{}}

\DeclareMathOperator*{\argmax}{arg\,max}
\newcommand{\norm}[1]{\left\lVert#1\right\rVert}

\usepackage[pagebackref=true,breaklinks=true,letterpaper=true,colorlinks,bookmarks=false]{hyperref}

\cvprfinalcopy 

\def\cvprPaperID{3740}

\newcommand{\bb}[1]{\bm{\mathrm{#1}}}

\ifcvprfinal\fi
\begin{document}


\title{
Self-supervised Learning of Dense Shape Correspondence 
}

\author{Oshri Halimi\\
Technion, Israel\\
{\tt\small oshri.halimi@gmail.com}
\and
Or Litany\\
Facebook AI Research\\
{\tt\small orlitany@fb.com}
\and
Emanuele Rodol\`{a}\\
Sapienza University of Rome\\
{\tt\small rodola@di.uniroma1.it}
\and
Alex Bronstein\\
Technion, Israel\\
{\tt\small bron@cs.technion.ac.il}
\and
Ron Kimmel\\
Technion, Israel\\
{\tt\small ron@cs.technion.ac.il}
}

\maketitle

\begin{abstract}
We introduce the first completely unsupervised correspondence learning approach for deformable 3D shapes. 
Key to our model is the understanding that natural deformations (such as changes in pose) approximately preserve the metric structure of the surface, yielding a natural criterion to drive the learning process toward distortion-minimizing predictions. 
On this basis, we overcome the need for annotated data and replace it by a purely geometric criterion. 
The resulting learning model is class-agnostic, and is able to leverage any type of deformable geometric data for the training phase. 
In contrast to existing supervised approaches which specialize on the class seen at training time, we demonstrate stronger generalization as well as applicability to a variety of challenging settings. 
We showcase our method on a wide selection of correspondence benchmarks, where we outperform other methods in terms of accuracy, generalization, and efficiency.
\end{abstract}


\section{Introduction}

\comment{
At the heart of virtual reality and next-generation 3D applications is the modeling of a virtual world to which the user is exposed.
In this context, a basic modeling feature is shape deformability, an inherent characteristic of the real world that should be part of any reasonable design of a virtual reality.
}

\comment{
In our era of virtual and augmented reality it is natural to expect that 3D models will be as common as images and videos. Data driven approach can replace traditional approach. 
}

\begin{figure}[t!]
    \begin{overpic}
    [trim=0cm 0cm 0cm 0cm,clip,width=1.0\linewidth]{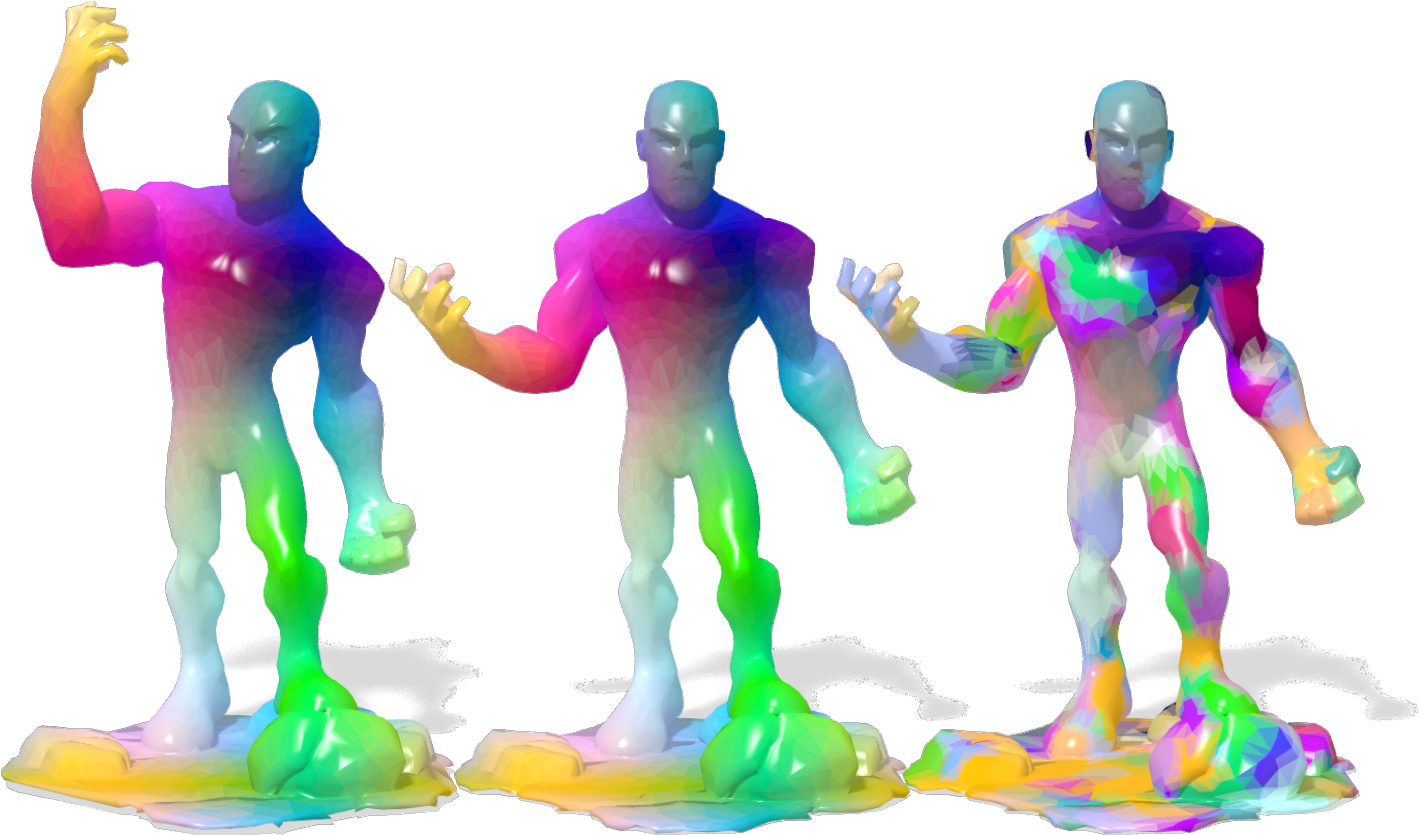}
    \put(38,56.8){\footnotesize Unsupervised}
    \put(39.7,54){\footnotesize (this paper)}
    \put(69,55){\footnotesize Supervised~\cite{litany2017deep}}
    \end{overpic}
\caption{\label{fig:teaser}
  Dense correspondence between articulated objects obtained with the proposed unsupervised loss.
  The network was optimized on a single (unlabeled) example. For comparison, we show the result of a state-of-the-art {\em supervised} network pre-trained on human shape. See Section \ref{sec:zeroshot} for more details. Correspondence is visualized by colors mapped from the leftmost shape.
  }
\end{figure}

The problem of finding accurate dense correspondence between non-rigid shapes is fundamental in geometry processing. 
It is a key component in applications such as deformation modeling, cross-shape texture mapping, pose and animation transfer to name just a few. 
Dense deformable shape correspondence algorithms can be broadly categorized into two families. 
The first can be referred to as \emph{axiomatic} or \emph{model-based}: 
A certain geometric assumption is asserted and pursuit for by some numerical scheme. 
Modeling assumptions attempt to characterize the action of a class of deformation on some geometric quantities commonly referred to as \emph{descriptors}. 
Such geometric quantities often encode local geometric information in the vicinity of a point on the shape (point-wise descriptors) such as normal orientation \cite{SHOT}, curvature \cite{pottmann09}, and heat \cite{sun2009concise} or wave \cite{aubry2011wave} propagation properties. Another type of geometric quantities are the global relations between pairs of points (pair-wise descriptors), which include geodesic \cite{elad2003bending,bronstein2006generalized}, diffusion \cite{Coifman7426} or commute time \cite{von2014hitting} distances.  
Given a pair of shapes, a dense map between them is sought to minimize the discrepancy between such descriptors. While the minimization of the point-wise discrepancies can be formulated as a linear assignment problem (LAP) and solved efficiently for reasonable scales, the use of pair-wise descriptors leads to a quadratic assignment problem (QAP) that is unsolvable for any practical scales. Numerous approximations and heuristics have been developed in the literature to alleviate the computational demand of QAPs. 

The second family of correspondence algorithms is \emph{data-driven} and takes advantage of modern efficient machine learning tools. 
Instead of axiomatically modeling the class of deformations and the geometric properties of the shapes of interest, these methods infer such properties from the data themselves. 
Among such approaches are learnable generalizations of the heat kernel signature \cite{litman2014learning}, as well as works interpreting correspondence as a labeling problem \cite{rodola2014dense}. Other recent methods generalize CNNs to non-Euclidean structures for learning improved descriptors \cite{monet, boscaini2016learning}. A recent method based on extrinsic deformation of a null-shape was introduced in \cite{groueix2018b}.  
A common denominator of these approaches is the \emph{supervised} training regime -- they all rely on examples of ground truth correspondences between exemplar shapes.  

A major drawback of this supervised setting is the fact that in the case of 3D shape correspondence the ground truth data are scarce and very expensive to obtain.
For example, despite being restricted to a single shape class (human bodies), the MPI FAUST scanning and labeling system \cite{dfaust:CVPR:2017} required substantial manual labor and considerable financial costs. 
In practice, labeled models are expected to be just a small fraction of the existing geometric data, bringing into question the scalability of any supervised learning algorithm.

\comment{
The effort involved in annotating such data could make sense at the object level, or maybe parts level. 
But, annotating at a high resolution of point to point matching is  too expensive and actually impractical; this calls for the design of techniques that would allow less supervision.
 We propose an unsupervised scheme to learn shape correspondence which does not rely on ground-truth labeling but rather on few axioms that encapsulate our knowledge about the nature of an admittable correspondence. 
 Our adopted axioms reflect the expectation that the desired correspondence preserve the metric and topological properties between the deformed shapes \cite{elad2003bending,bronstein-face}. 
 A given correspondence that consistently satisfies these axioms is expected to be the true correspondence. 
 While such axioms are conceptually easy to formulate, it is generally hard to find the correspondence that fulfills them using traditional optimization techniques. 
 The power of a deep learning approach is revealed as being a powerful optimization tool. 
 In our case we show how axioms replace the need of external ground truth data. 
 
 The proposed method is expected to work as long as we can construct a model that describes the properties of the desired correspondence.
As an example, in this paper, we suggest to treat shapes of the same subject under various poses as being isometrically similar. 

}

\subsection{Contribution}

We propose an unsupervised learning scheme for dense 3D deformable shape correspondence based on a purely geometric criterion. 
The suggested approach bridges between the model-based and the data-driven worlds by learning point-wise descriptors that result in correspondences minimizing pair-wise geodesic distance disagreement. 
The correspondence is then solved for using the functional maps framework \cite{ovsjanikov12} totally avoiding the computational burden of the pair-wise methods. 
While the point-wise descriptors are learned on a surrogate task only approximately characterizing the real data (which deviate from the asserted isometric deformation model), the method shows excellent generalization capabilities exceeding the supervised counterparts without ever seeing examples of ground truth correspondences. 
To the best of our knowledge, this is the first unsupervised approach applied to the geometric problem of finding shape correspondence.  

\comment{
Winning traditional algos - with respect to the performance and still generalizing (prove both claims with graphs)
Winning Learning algos - same performance, don't use data!!!! 

Integrating axiomatic reasoning about invariant properties of shapes, like the inter-geodesic distances between surface points, allow us to efficiently train a deep neural network that lends itself for accurate matching of shapes. 
The suggested marriage between an axiomatic methodology within a deep neural network framework brings up the best of these seemingly unrelated two worlds.
If so Emphasize this? \orl{be careful here. Basically all traditional methods pre-deep learning are unsupervised...we're combining this domain knowledge with modern and more accurate DL techniques}
\oshri{Right! The idea is that we have two frontiers: traditional algos and learning algos. We want to show that we are superior in both frontiers. Traditional algos are indeed unsupervised, but have lesser performance. We win. Learning algos have good performance but use ground truth data. We don't, again we ain. Conclusion: the fusion of axioms + leaning is winning each of the paradigms.}
}

A major advantage of the proposed framework is when the data themselves are scarce, in extreme conditions we might have only one pair of shapes that we would like to match and we do not have a training dataset that contains similar shapes. 
While a supervised scheme depends on a relatively large amount of labelled data to deduce a generalizing model, with the unsupervised network we can simply optimize on a single pair of shapes that by itself contains two training samples, one in each direction of the correspondence.
Our experiments required only a few iterations that take just a couple of  minutes to run. 
Usually, less than $100$ iterations were more than enough. 
As a result we obtain an accurate matching between the shapes, see Figure \ref{fig:teaser}. 
For a trained network the inference phase takes less than a second. 
We believe that this strategy has its own merits as a replacement of the existing computationally expensive methods that are based on pair-wise descriptors. 
The framework can be interpreted as a fusion between the previously proposed FMNet, the Functional Maps centered network architecture \cite{litany2017deep} and the pair-wise geodesic distance distortion criterion used in previously proposed model-based approaches such as GMDS \cite{bronstein2006generalized}.

\section{Background}

\subsection{Minimum distortion correspondence}
We model shapes as Riemannian 2-manifolds $\mathcal{X}$ equipped with a distance function $d_\mathcal{X}:\mathcal{X}\times\mathcal{X}\to\mathbb{R}$ induced by the standard volume form. An {\em isometry} is a map $\pi:\mathcal{X}\to\mathcal{Y}$ satisfying, for any pair $x_1,x_2\in\mathcal{X}$:
\begin{equation}\label{eq:isometry}
    d_\mathcal{X}(x_1,x_2) = d_\mathcal{Y}({\pi}(x_1),{\pi}(x_2))\,.
\end{equation}
Correspondence seeking approaches optimize for a map ${\pi}$ satisfying the distance preservation criterion~\eqref{eq:isometry}. In practical applications, only approximate realizations of an isometry are expected; thus, one is interested in identifying a distortion-minimizing map of the form
\begin{equation}\label{eq:distortion}
    \pi^\ast = \arg\hspace{-0.1cm}\min_{\pi:\mathcal{X}\to\mathcal{Y}}
    \hspace{-0.2cm}
    \sum_{\substack{x_1,x_2 \in \mathcal{X}}}
    \hspace{-0.4cm}
    \left( d_\mathcal{X}(x_1,x_2) - d_\mathcal{Y}({\pi}(x_1),{\pi}(x_2)) \right)^2.
\end{equation}

In the discrete setting, we assume manifolds $\mathcal{X},\mathcal{Y}$ to be represented as triangle meshes sampled at $n$ vertices each. Minimum distortion correspondence thus takes the form of a quadratic assignment problem (QAP), where the minimum is sought over the space of $n \times n$ permutation matrices.

Several studies have tried to reduce the complexity of this QAP at the cost of getting an approximate solution via sub-sampling \cite{tevs2011intrinsic,rodola2012game}, hierarchical matching \cite{bronstein2006generalized,wang2011discrete} or convex relaxations \cite{aflalo2015convex, chen2015robust}. 

However, complicated to solve, the minimum distortion criterion \eqref{eq:distortion} is axiomatic and does not require any annotated correspondences, making it a natural candidate for an unsupervised learning loss.

\subsection{Descriptor learning}

A common way to make the optimization of \eqref{eq:distortion} more efficient is by restricting the feasible set to include only potential matches among points with similar descriptors. By doing so, one shifts the key difficulty from optimizing a highly non-linear objective to designing deformation-invariant local point descriptors. 

This has been an active research goal in shape analysis in the last few years, with examples including GPS~\cite{rustamov2007laplace}, heat and wave kernel signatures~\cite{sun2009concise,aubry2011wave}, and the more recent geodesic distance descriptors~\cite{shamai2017geodesic}. In 3D vision, several {\em rotation}-invariant geometric descriptors have been proposed~\cite{SHOT, johnson1999using}. Despite their lack of invariance to isometric deformations, however, the adoption of extrinsic descriptors has been advocated in deformable settings \cite{rodola2017partial} due to their locality and resilience to boundary effects.
Handcrafted descriptors suffer from an inherent drawback of requiring manual tuning. Learning techniques have thus been proposed to define descriptors whose invariance classes are learned from the data. Early examples include approaches based on decision forests and metric learning \cite{litman2014learning,rodola2014dense,cosmo2016matching}; more recently, several papers have proposed an adaptation of deep learning models to non-Euclidean domains, achieving dramatic improvement. 
In \cite{masci2015geodesic,boscaini2016learning,monet} learnable local filters were introduced based on the notion of patch operator. 

In~\cite{litany2017deep} a task driven approach was taken instead, where the network learns descriptors which excel at the task at hand in a supervised manner (this will be discussed in detail in Section~\ref{sec:fmnet}). As we will show in the sequel, our approach builds upon this model while completely removing the need for supervision.

\subsection{Functional maps}
The notion of {\em functional map} was introduced in \cite{ovsjanikov12} as a tool for transferring functions between surfaces without the direct manipulation of a point-to-point correspondence. 
Let $\mathcal{F}(\mathcal{X}),\mathcal{F}(\mathcal{Y})$ be real-valued functional spaces defined on top of $\mathcal{X}$ and $\mathcal{Y}$ respectively. Then, given a bijection $\pi:\mathcal{X}\to\mathcal{Y}$, the functional map $T:\mathcal{F}(\mathcal{X})\to\mathcal{F}(\mathcal{Y})$ is a linear mapping acting as
\begin{equation}
    T(f) = f \circ \pi^{-1} \,.
\end{equation}
The functional map $T$ admits a matrix representation with respect to orthogonal bases $\{\phi_i\}_{i\ge 1},\{\psi_i\}_{j\ge 1}$ on $\mathcal{X}$ and $\mathcal{Y}$ respectively, with coefficients $\textbf{C}=(c_{ij})$ calculated as follows:
\begin{equation}
T(f) = \sum_{ij} \langle \phi_i ,f\rangle \underbrace{\langle T \phi_i , \psi_j \rangle}_{c_{ji}}\psi_j\,.
\end{equation}

While the functional maps formalism makes no further requirements on the chosen bases, a typical choice is the Laplace-Beltrami eigenbasis (the justification for the optimality of this choice can be found in
\cite{AflBreKim15}). 

Truncating these series to $k$ coefficients, one obtains a band-limited approximation of the functional correspondence $T$. Specifically, the map
\begin{equation}
P : x \mapsto \sum\limits_{i,j} c_{ji} \phi_i(x) \psi_j\,,
\end{equation}
also referred to as a \emph{soft map}, will assign to each point $x\in\mathcal{X}$ a function concentrated around $y=\pi(x)$ with some spread. 

To solve for the matrix $\bb{C}$, linear constrains are derived from the knowledge of knowingly corresponding functions on the two surfaces. Corresponding functions are functions that preserve their value under the mapping $T$. 
Given a pair of corresponding functions $f:\mathcal{X} \rightarrow \mathbb{R}$ and $g:\mathcal{Y} \rightarrow \mathbb{R}$ with  coefficients $\bb{\hat{f}} = \{ \langle \phi_i,f\rangle \}_i$ and $\bb{\hat{g}} = \{ \langle \psi_j,g\rangle \}_j$ in the bases $\{\phi_i \}$ and  $\{ \psi_j \}$ respectively, 
the correspondence imposes the following linear constraint on $\bb{C}$
\begin{equation}
	\bb{\hat{g}} = \bb{C} \bb{\hat{f}}.
\end{equation}
Each pair of such corresponding functions is translated into a linear constraint.
 
Suppose there exists an operator receiving a shape $\mathcal{X}$ and producing a set of \emph{descriptor functions} on it. Let us further assume that given another shape $\mathcal{Y}$, the operator will produce a set of corresponding functions related by the latent correspondence between $\mathcal{X}$ and $\mathcal{Y}$. 
In other words, applying the above operator on the said pair of shapes produces a set of pairs of corresponding functions $(f_i,g_i)$, each pair comprising  $f_i$ defined on $\mathcal{X}$ and $g_i$ on $\mathcal{Y}$. We stack the corresponding coefficients $\bb{\hat{f}}_i$ and $\bb{\hat{g}}_i$ into the columns of the matrices $\bb{\hat{F}}$ and $\bb{\hat{G}}$. 
The functional map matrix $\bb{C}$ is then given by the (least squares, or otherwise regularized) solution to the system
\begin{equation}
\bb{\hat{G}} = \bb{C} \bb{\hat{F}}.
\label{opt}
\end{equation}
Thus, the requirement for specific knowledge of the point-to-point correspondence is replaced by the relaxed requirement of knowledge about functional correspondence.

\subsection{Deep functional maps}\label{sec:fmnet}
A significant caveat in the above setting is that, unless the shapes $\mathcal{X}$ and $\mathcal{Y}$ are related by a narrow class of deformations, it is very difficult to construct an operator producing a sufficient quantity of stable and repeatable descriptors. However, such an operator can be \emph{learned} from examples. The aim of the deep functional maps network (FMNet) introduced in \cite{litany2017deep} was to learn descriptors which, when used in the above system of equations, will induce an accurate correspondence. At training time, FMNet operates on input descriptor functions (e.g. SHOT descriptors), and improves upon them by minimizing a geometric loss that is defined on the soft correspondence derived from the functional map matrix. The differentiable functional map layer (FM), solves the equation (\ref{opt}), with the current descriptor functions in each iteration.

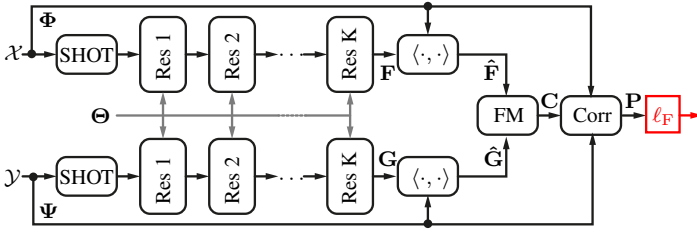
\begin{figure}[h]
\centering
\begin{overpic}
		[trim=0cm 0cm 0cm 0cm,clip,width=1.1\linewidth]{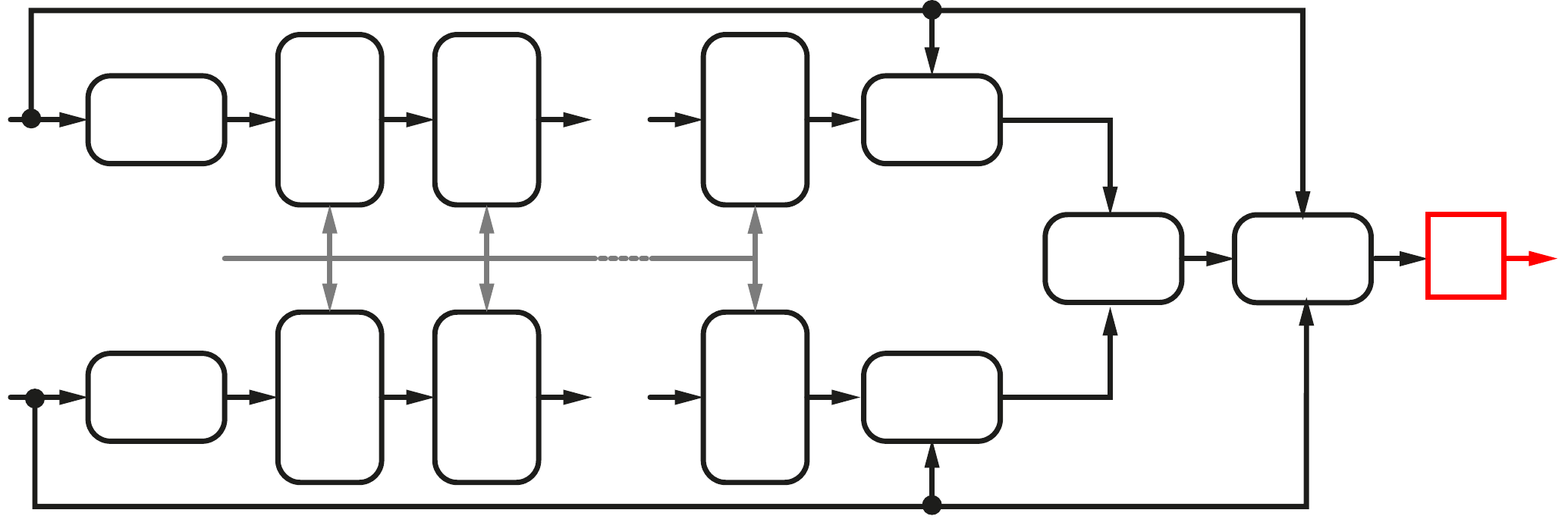}
		\put(-2,25.4){\footnotesize $\mathcal{X}$}
		\put(3,30.2){\footnotesize $\bb{\Phi}$}
		\put(6,25.2){\footnotesize SHOT}
		\put(20,22.5){\footnotesize \rotatebox{90}{Res 1}}
		\put(30,22.5){\footnotesize \rotatebox{90}{Res 2}}
		\put(47,22.5){\footnotesize \rotatebox{90}{Res K}}
		\put(37.8,25.4){\footnotesize $\cdots$}
		\put(57,25.5){\footnotesize $\langle \cdot , \cdot \rangle $}
		\put(67.5,22.6){\footnotesize $\bb{\hat{F}}$}
		\put(52.5,22.6){\footnotesize $\bb{F}$}

		\put(10.5,16.3){\footnotesize $\bb{\Theta}$}		
		\put(69,16.5){\footnotesize FM}

		\put(75.7,18.5){\footnotesize $\bb{C}$}
		\put(88,18.5){\footnotesize $\bb{P}$}

		\put(80.5,16.5){\footnotesize Corr}
		\put(92,16.5){\footnotesize {\color{red} $\ell_{\mathrm{F}}$ } }

		\put(3,2.5){\footnotesize $\bb{\Psi}$}
		\put(-2,7.5){\footnotesize $\mathcal{Y}$}
		\put(6,7.5){\footnotesize SHOT}
		\put(20,4.5){\footnotesize \rotatebox{90}{Res 1}}
		\put(30,4.5){\footnotesize \rotatebox{90}{Res 2}}
		\put(47,4.5){\footnotesize \rotatebox{90}{Res K}}
		\put(37.8,7.4){\footnotesize $\cdots$}
		\put(57,7.5){\footnotesize $\langle \cdot , \cdot \rangle $}
		\put(67.5,10){\footnotesize $\bb{\hat{G}}$}
		\put(52.2,10){\footnotesize $\bb{G}$}
\end{overpic} \caption{\label{fig:fmnet}Deep Functional Maps network architecture \cite{litany2017deep}}
\end{figure}
The network architecture described in \cite{litany2017deep} consists of $7$ fully-connected residual layers with exponential linear units (ELU) and no dimentionality reduction. 
The output of the residual network is a dense vector-valued descriptor. Given two shapes  $\mathcal{X}$ and $\mathcal{Y}$, the descriptors are calculated on each shape using the same network, and are projected onto the corresponding truncated LBO bases. 
The resulting coefficients are given as an input to the functional map (FM) layer that calculates the functional map matrix $\bb{C} \in \mathbb{R}^{k \times k}$ according to (\ref{opt}). The following correspondence layer (Corr) produces a soft correspondence matrix $\bb{P} \in \mathbb{R}^{n_\mathcal{Y} \times n_\mathcal{X}}$ out of the functional map matrix $\bb{C}$,
\begin{equation}
    \label{soft}
    \bb{P} = |\bb{\Psi} \bb{C} \bb{\Phi}^{T} \bb{A} |_{\| \cdot \|}.
\end{equation}
Where we denoted the number of vertices on the discretized shapes as ${n_\mathcal{X}}$ and ${n_\mathcal{Y}}$, and the diagonal matrix $\bb{A}$ normalizes the inner products with the discrete area elements of $\mathcal{X}$. 
The absolute value and the $L_2$ column normalization, denoted by $_{\| \cdot \|}$, ensure that the values of $p_{ji}^2$ can be interpreted as the probability of vertex $j$ on shape $\mathcal{Y}$ being in correspondence with vertex $i$ on $\mathcal{X}$. We denote the element-wise square of $\bb{P}$ by $\bb{Q} = \bb{P} \circ \bb{P}$, with $\circ$ standing for the Hadamard product.


%
Treating the $i$-th column of $\bb{Q}$, $\bb{q_i}$, as the distribution on the points of $\mathcal{Y}$ corresponding to the point $i$ on $\mathcal{X}$, we can evaluate the expected deviation from the ground truth correspondence $\pi^*(i)$. This is expressed by the second-order moment 
\begin{eqnarray}
\mathbb{E}_{j \sim \bb{q_i} }  d^2_{\mathcal{Y}}(j,\pi^*(i)) = \sum_{j \in \mathcal{Y}}q_{ji} d^2_{\mathcal{Y}}(j,\pi^{*}(i)).
\end{eqnarray}
where  $d_{\mathcal{Y}}(j,\pi^{*}(i))$ is the geodesic distance on $\mathcal{Y}$ between the vertex $j$ and the ground truth match $\pi^{*}(i)$ of the vertex $i$ on $\mathcal{X}$.
As usual, this moment comprises a variance and a bias terms; while the former is the result of the band-limited approximation (due to the truncation of the basis), the latter can be controlled. 

Averaging the above moment over all points on $\mathcal{X}$ leads to the following supervised loss
\begin{eqnarray}
    \label{loss_supervised}
    \ell_{\mathrm{sup}} (\mathcal{X},\mathcal{Y}) &=&  \frac{1}{|\mathcal{X}|} \sum_{i \in \mathcal{X}}\sum_{j \in \mathcal{Y}}q_{ji} d^2_{\mathcal{Y}}(j,\pi^{*}(i)) \nonumber\\
    &=& \frac{1}{|\mathcal{X}|} \norm{\bb{P} \circ (\bb{D}_{\mathcal{Y}} \bb{\Pi}^* ) }_\mathrm{F}^2,
\end{eqnarray}
where $\bb{D}_{\mathcal{Y}}$ denotes the pairwise geodesic distance matrix evaluated for each shape at the pre-processing stage, and $\bb{\Pi}^*$ is the ground truth permutation relating between the shapes.  The batch loss is the sum of $\ell_{\mathrm{sup}}(\mathcal{X},\mathcal{Y})$ for all the pairs in the minibatch. 

Training an FMNet follows the standard Siamese setting commonly used for descriptor or metric learning, in which two copies of the network with shared parameters produces the descriptors on $\mathcal{X}$ and $\mathcal{Y}$. From this perspective, the functional map and the soft correspondence layers are parts of the Siamese loss rather than of the network itself.

\section{Unsupervised deep functional maps}
The authors in \cite{litany2017deep} showed that FMNet achieves state-of-the-art performance on standard deformable shape correspondence benchmarks. 
However, one can argue that the supervised training regime is prohibitive in terms of the amounts of the manually annotated data required.
\comment{Even when pairs of shapes with dense ground truth correspondence come from an automatic alignment process, such as in the FAUST dataset \cite{faust}, it introduces modeling bias -- in the case of FAUST, it is the common connectivity of the fitted model. }

The main contribution of this paper is the transition to an unsupervised training regime, i.e., a setting requiring no ground-truth correspondence. 
The key idea is that even if ground truth correspondence is not provided, we can still evaluate the quality of the resulting correspondence based on the preservation of standard geometric quantities.
As mentioned before, human pose articulation can be modeled as approximate isometries, that is, the latent correspondence introduces little metric distortion. 
If two vertices were at some geodesic distance on the source shape, after mapping by the correct correspondence, the distance between corresponding points on the target domain is preserved. 

Let $\bb{P}$ be the output of the soft correspondence layer of an FMNet; as before, its squared elements $q_{ji} = p_{ji}^2$ are interpreted as probability distributions on $\mathcal{Y}$. In these terms, the $ji$-th element of the matrix $\bb{Q}^T \bb{D}_{\mathcal{Y}} \bb{Q}$ 
\begin{equation}
    (\bb{Q}^T \bb{D}_{\mathcal{Y}} \bb{Q})_{ji} = \sum_{m,n} p_{mi}^2 p_{nj}^2 d_{\mathcal{Y}}(m,n) 
\end{equation}
represents the expected distance on $\mathcal{Y}$ between the images of the vertices $i,j \in \mathcal{X}$ under the soft correspondence $\bb{P}$. 

This allows to define the following \emph{unsupervised} loss
\begin{equation}
    \label{unsupervised_loss}
    \ell_{\mathrm{uns}} (\mathcal{X},\mathcal{Y}) =\frac{1}{|\mathcal{X}|^2}  \norm{\bb{D}_{\mathcal{X}} - \bb{Q}^T \bb{D}_{\mathcal{Y}} \bb{Q}   }^2_{\mathrm{F}}.
\end{equation}
 The batch loss is the sum of $\ell_{\mathrm{uns}}(\mathcal{X},\mathcal{Y})$ for all the pairs in the minibatch. 
This loss measures the $L_2$ geodesic distance distortion and can be interpreted as a soft correspondence version of the GMDS loss, see  \cite{aflalo2016spectral}.
Note that rather than solving the QAP directly, we propose to train an FMNet using $\ell_{\mathrm{uns}}$, which promotes the network to generate descriptors for which the resulting soft correspondence minimizes the expected pairwise distance distortion.  

From the point of view of the unsupervised network all the shapes in the world could constitute a training set. 
Since the network does not use any ground-truth correspondence data, its learning is not limited to a class of shapes and is expected to improve when new shapes are encountered. 
The strict separation between the training and test sets that characterizes the supervised regime does not strictly apply in the unsupervised setting, since the network does not make any distinction between training and testing shapes. 
If the training set is representative enough to generalize the test set, the network can learn on the training set and infer on the test set, reducing the processing time per shape. 
Contrarily, if the training set is not representative enough, the network can still gain advantage from being exposed to the test shapes and using them to improve the learned model.
Learning could in fact be executed even at inference time. 
For the FAUST scans \cite{faust}, for example, the authors provide a training test with ground-truth labeling and a disjoint test set without the labels. 
A supervised scheme cannot gain much 
from seeing the test shapes since they lack the ground truth correspondence, and it is confined to train on the training set alone. 
The unsupervised scheme has access to the same data, but contrarily to the supervised counterpart, it can use the unlabeled test shapes to improve prediction accuracy. 
\comment{Due to this reason, for the FAUST benchmark we report two scores, the first one following the traditional supervised setting, in which separation is made to training and test sets, and only the training set is used for training. 
Additionally, we report a second score achieved by injecting a subset of the test set to the training process. 
The injected shapes are not included in the specific test challenge pairs list. 
We show that by encountering additional unsupervised data the network can improve its prediction. }

\section{Implementation}
We implemented our network in TensorFlow \cite{tensorflow2015-whitepaper}, running on a GeForce GTX 1080 Ti GPU. Data preprocessing and correspondence refinement were done in Matlab. 
\subsection{Pre-processing}
To enable mini-batches of multiple shapes to fit in memory, each shape in the
training set was remeshed to between $n \sim 5$K and $7$K vertices , by edge contraction \cite{garland1997surface}. For each remeshed shape $k \sim 70 - 150$ Laplace-Beltrami eigenfunctions were calculated as well as a $352$-dimensional SHOT descriptor~\cite{salti2014shot}, using 10 bins and a SHOT radius that was roughly chosen to 5\% of the maximal pairwise geodesic distance. Geodesic distance matrices $\bb{D}$ were estimated using the fast marching method~\cite{kimmel1998computing}. These quantities constitute the input to the network.

\subsection{Network architecture and loss}
For a more direct and fair comparison, we adopted the same network architecture as FMNet presented in Figure \ref{fig:fmnet}. Specifically, the input for each pair of shapes is their $n\times k$ truncated LBO bases $\bb{\Phi}$ and $\bb{\Psi}$ , the $n\times n$ pairwise distance matrices $\bb{D}_\mathcal{X}$ and $\bb{D}_\mathcal{Y}$,
and the $n\times 352$ SHOT descriptor fields $\bb{S}_\mathcal{X},\bb{S}_\mathcal{Y}$. These are fed to a $7$-layer residual network~\cite{he2016deep} outputting $352$-dimensional dense descriptor fields $\bb{F}$ and $\bb{G}$ on $\mathcal{X}$ and $\mathcal{Y}$ respectively, which can be thought of as non-linearly transformed variants of SHOT.

The computed descriptors are then input to the functional map layer, yielding a functional map matrix $\bb{C}$ according to \eqref{opt}, followed by a soft correspondence layer producing the stochastic correspondence matrix $\bb{P}$ as per Eq.~\eqref{soft}. Finally, the unsupervised loss is calculated according to Eq.~\eqref{unsupervised_loss}.
While in FMNet the loss is calculated on a random sub-sampling of the vertices, we found that this strategy introduces inaccuracies to the descriptor coefficients in the LBO basis. When sub-sampling is used, the network is only able to evaluate an {\em estimate} of the projection of the descriptors onto the LBO basis, which quickly becomes inaccurate for descriptors with high-frequency content. To avoid this, in our implementation we perform the projection at full resolution while decreasing the size of the mini-batch to $\sim$4--5 pairs of shapes per mini-batch. In all our experiments we used no more than a few thousands (about 3K--10K) mini-batch iterations. For comparison, the supervised FMNet used 100K iterations of $32$ mini-batch size to achieve similar results.
\subsection{Post-processing}
\label{sec:postprocess}
\vspace{1ex}\noindent\textbf{Point-wise map recovery.}
Following the protocol of FMNet, we apply the product manifold filter (PMF)~\cite{vestner2017product} to improve the raw prediction of the network. This algorithm takes noisy matches as input, and produces a (guaranteed) bijective and smoother correspondence of higher accuracy as output. 
The application of PMF boils down to solving a series of linear assignment problems $\argmax_{\bm{\Pi}^t}\langle \bm{\Pi}^t, \mathbf{K}_\mathcal{X}\bm{\Pi}^{t-1}\mathbf{K}_\mathcal{Y}^\top \rangle_{F}$, where $\bm{\Pi}^t$ ranges over the space of permutations, and $\mathbf{K}_\mathcal{X},\mathbf{K}_\mathcal{Y}^\top$ are kernel matrices acting as a diffusion operators. We refer to~\cite{vestner2017product} for additional details.

\vspace{1ex}\noindent\textbf{Upscaling.}
Since we operate on remeshed shapes, we finally apply an upscaling step to bring the correspondence back to the original resolution. Again, we follow the procedure described in FMNet~\cite{litany2017deep}, namely we solve a functional map estimation problem of the form
\begin{equation}
    \mathbf{C}_\mathrm{up} = \arg\min_\mathbf{C} \| \mathbf{C}\hat{\mathbf{F}}_\mathrm{up} - \hat{\mathbf{G}}_\mathrm{up}\|_{2,1}\,,
\end{equation}
where $\hat{\mathbf{F}}_\mathrm{up},\hat{\mathbf{G}}_\mathrm{up}$ contain the LBO coefficients (in the full resolution basis) of delta functions supported at corresponding points, extracted from the low resolution map $\mathbf{C}$. The $\ell_{2,1}$ norm (defined as the sum of $\ell_2$ norms of the columns) allows to downweight potential mismatches.

\section{Experiments}
\subsection{Learning to match a single pair}
\label{sec:zeroshot}
Before delving into training on large datasets, we begin our experimental section with testing one extreme of the shape matching problem: single input pair. Clearly, this is the native environment for classical, non-learning based methods. While learning based methods have endowed us with better solutions given large train sets, they are not equipped to handle entirely novel examples. Having developed an unsupervised network we demonstrate it can be utilized as an ad-hoc solver for a single pair, producing excellent results. In Figure \ref{fig:teaser} we show our result on a pair of shapes made by an artist\footnote{credit to the artist appears in the Acknowledgements section}. Note that we do not have groundtruth correspondences for this pair and therefore a supervised learning based method cannot be fine-tuned on the input pair. Instead we compare with raw predictions of FMNet that has been pre-trained on human shapes from FAUST. In addition, we ran the post processing procedure described in \ref{sec:postprocess}. While we got comparable results to our method (please see Appendix \ref{app:scarce} for visualization), runtime has exceeded one hour. Conversely, optimizing our network took about $15$ minutes. Furthermore, had we been given an additional deformation of the same shape to solve for, an axiom-based method would have needed to solve the problem from scratch. Differently, as our method had already learned to convert the pair-wise optimization problem to a descriptor matching problem, inference would take about one second!

\subsection{Faust synthetic}
Faust synthetic models \cite{faust} are a widely used data in shape matching tasks. In this experiment we use it for comparing our unsupervised method and its supervised counterpart under the same setting. We show that (a) optimizing for the unsupervised loss results in a correlated decrease of the supervised loss; (b) the  unsupervised method, achieves the same accuracy as the supervised one.
For training our network, We followed the same dataset splits as in \cite{litany2017deep} where the first $80$ shapes of $8$ subjects are used for training, while a disjoint set of $20$ shapes of $2$ other subjects were used for testing. Each training mini-batch contained $4$ pairs of shapes in their full resolution of $6890$ vertices. Since the corresponding vertices in the dataset have corresponding indices, we shuffled randomly the vertices of each shape in the training pair, for every new appearance in the training mini-batches, creating a nontrivial permutation between the corresponding vertices. This step is intended to eliminate the possibility that the network converges to descriptors that lead always to the trivial permutation. We used the same parameters as in \cite{litany2017deep}, namely, $k=120$ eigenfunctions and ADAM optimizer with a learning rate of $\alpha = 10^{-3}, \beta_1 = 0.9, \beta_2 = 0.999$ and $\epsilon = 10^{-8}$. We used $3K$ training mini-batches. Note that, as in \cite{litany2017deep}, since we train on pairs of shapes we have an effective train set size of $6400$.

\begin{figure}
    \includegraphics[width=\linewidth]{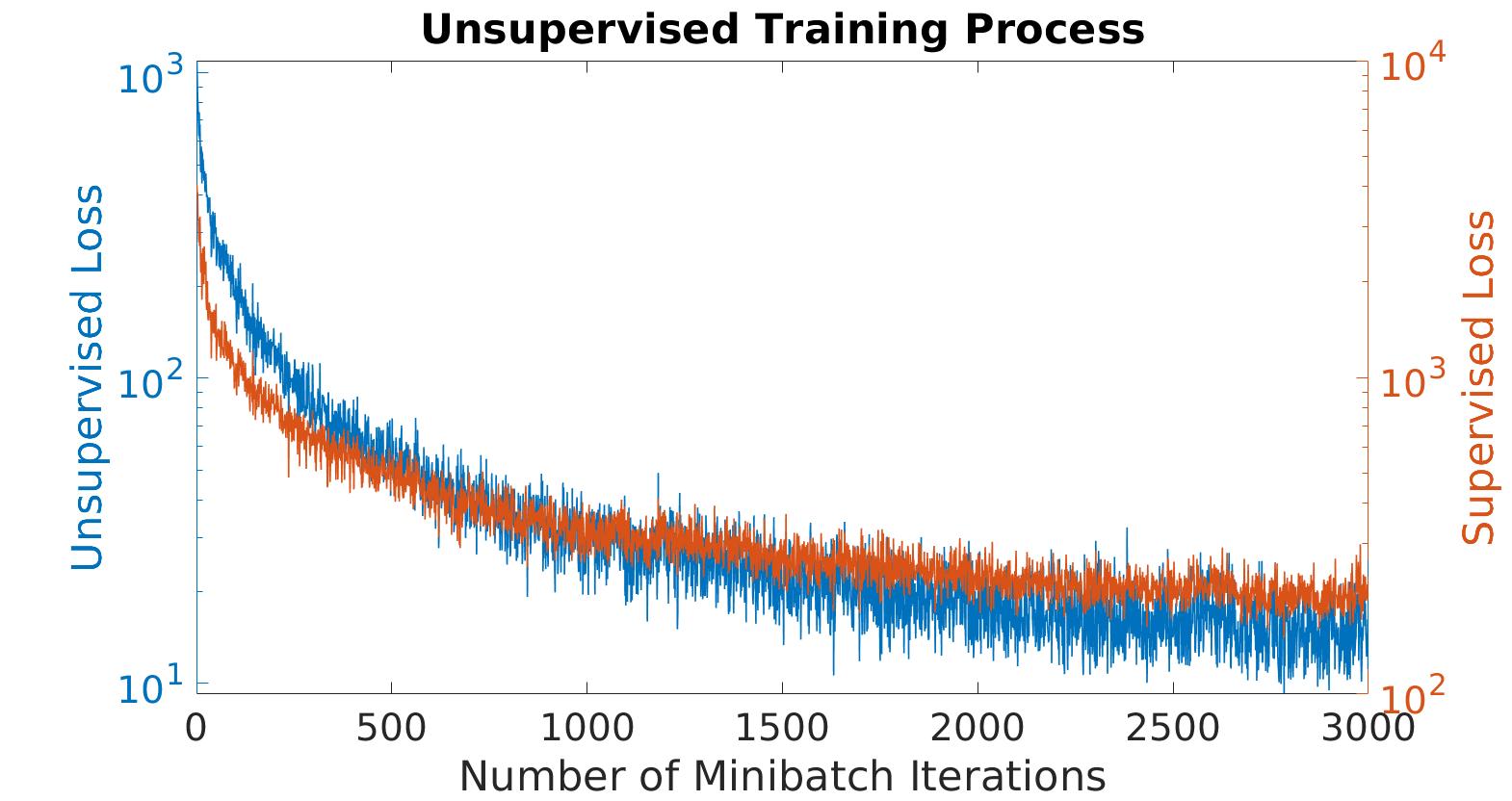}
    \caption{Unsupervised loss (left axis) and supervised loss (right axis) measured during the unsupervised training process, in logarithmic scale. While training target is the unsupervised loss, the supervised loss is decreased as a by-product.}
    \label{fig:anal}
\end{figure}

\vspace{1ex}\noindent\textbf{Loss function analysis.}
Figure \ref{fig:anal} displays the unsupervised loss during the training process (top), alongside with the supervised loss (bottom). Importantly, the unsupervised network had no access to ground truth correspondence.  

From the graphs, it can be observed that while the optimization target is the unsupervised loss, the supervised loss is decreased as well. This demonstrates nicely that when our underlying assumption of (quasi-) isometric deformations holds, one can replace the expensive supervision altogether with a single axiomatic-driven loss term. 

\begin{figure}

    \includegraphics[width=\linewidth]{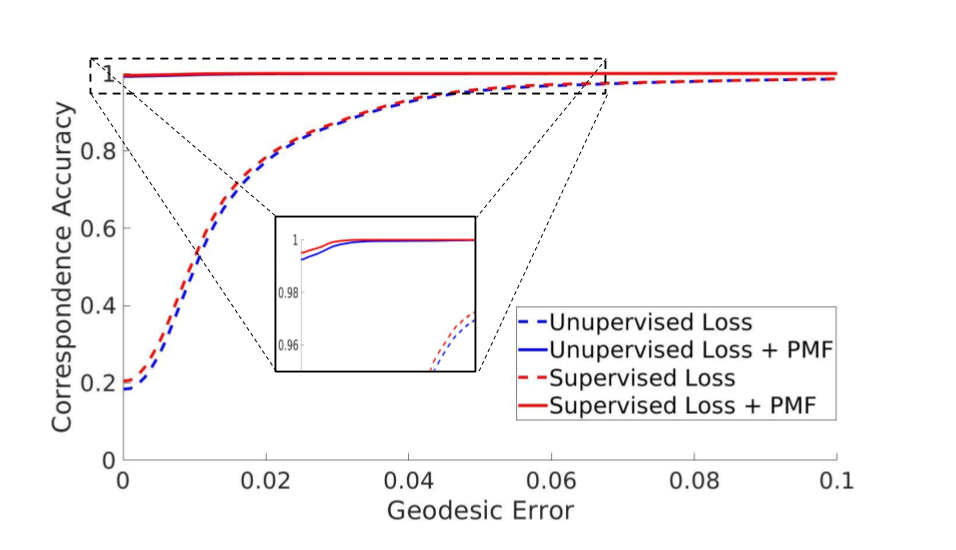}
    \caption{\label{isocomp} Unsupervised and supervised network results, evaluated on synthetic Faust intra-subject pairs. Performance is practically the same, and we zoom in to show the separate curves.}
\end{figure}

\begin{figure}
    \includegraphics[width=1.1\linewidth]{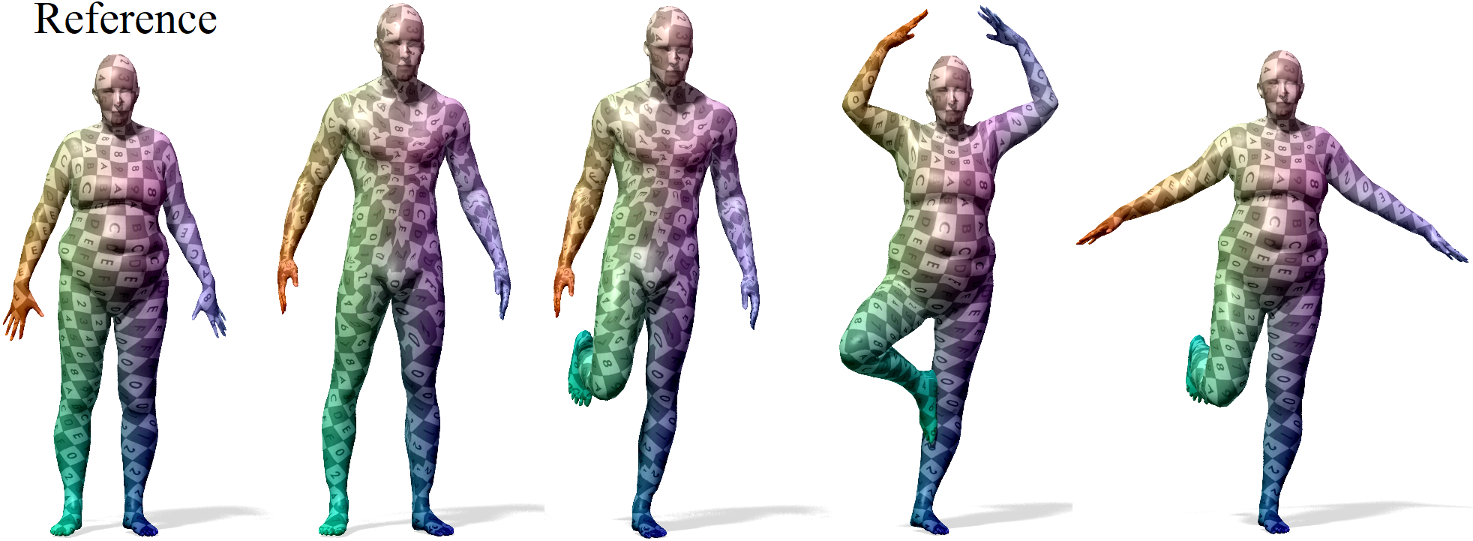}
    \caption{\label{synthetic_vis}Synthetic Faust texture transfer. Left woman - reference model. Four right models show the predicted matching.}
\end{figure}

\vspace{1ex}\noindent\textbf{Performance comparison.}
To compare our results with the supervised network, we followed the same training scheme, this time using the supervised loss. We used the $20$ test shapes to construct a test dataset of $400$ pairs in total; $200$ of which are of the same subject at different poses, and the other $200$ are of different subjects at different poses (Note that the matching is directional from source to target, hence this set is not redundant). The intra-subject pairs, are well modeled by isometry while the inter-subject pairs exhibit deviation from isometry. For each test pairs the vertices of both shapes were shuffled separately in a random manner (but consistent between the two networks).  
Figure \ref{isocomp} compares the results for the $200$ intra-subject pairs in synthetic Faust. Figure \ref{synthetic_vis} visualizes the calculated correspondences.

\subsection{Real scans}
Traditionally, axiom-based methods were proven useful only in the Computer Graphics regime. One of our goals in introducing learned descriptors is to demonstrate the applicability of our method to real scanned data. To this end, we make use of FAUST real scans benchmark. These are very high-resolution, non-watertight meshes, many of which contain holes and topological noise.
We used the dataset split as prescribed in the benchmark. The scans were down-sampled to a resolution of $7$K vertices. For each scan the distance matrix was calculated, as well as $352$-dimensional SHOT descriptors and $k=70$ LBO eigenfunctions. Each training mini-batch contained $4$ pairs of shapes. We trained our network for $10K$ iterations. The raw network predictions were only upscaled but not refined with PMF. \comment{Several qualitative results are provided in Figure \ref{fig:faust_scans}.}Quantitative results were evaluated through the online evaluation system. With an average and worst case scores of $\bb{2.51}$ cm and \textbf{$\bb{24.35}$} cm, respectively,  on the intra challenge, our network performs on par with state of the art methods that do not use additional data; namely, FMNet ($2.44$, $26.16$), and Chen et al.~\cite{chen2015robust} ($4.86$, $26.57$). We perform slightly below the recent 3D-CODED method  \cite{groueix2018b} ($1.98$, $5.18$) which uses an additional augmentation of over $200$K shapes at training. The same method, when not using additional data achieves worse results by a factor of $\approx 9$. 

\comment{\oshri{Practically the bottleneck for ultimate results on FAUST scans was the topological changes. I can make an experiment with diffusion distances for the supplementary}
\orl{We should definitely do it. But as far as I know, we can't show *new* results in the supp.}}
\comment{
\begin{figure}
    \includegraphics[width=0.5\linewidth]{scans_final.png}
    \caption{\label{fig:faust_scans}Visualization of FAUST scans test results \oshri{placeholder}}
\end{figure}
}
\subsection{Generalization}
\begin{figure}
\begin{center}
    \input{./tosca.tikz}
    \hspace{0.1cm}
    \begin{overpic}
    [trim=0cm 0cm 0cm 0cm,clip,width=0.35\linewidth]{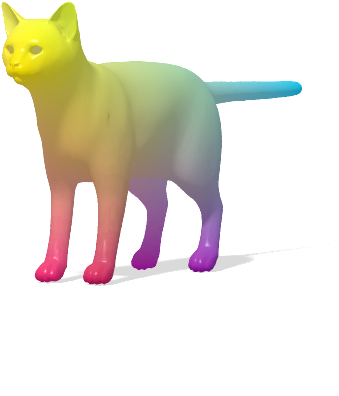}
    \put(14,105){\footnotesize Reference}
    \end{overpic}
    \begin{overpic}
    [trim=0cm 0cm 0cm 0cm,clip,width=1.0\linewidth]{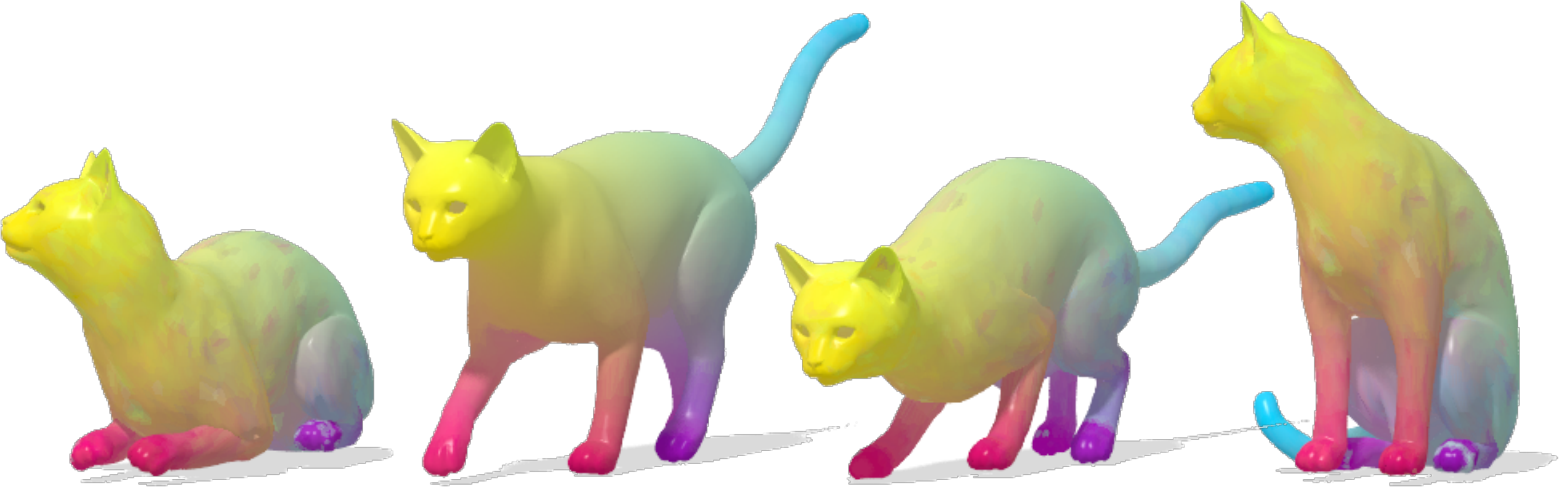}
    \end{overpic}
    \caption{\label{fig:tosca}Generalization experiment on TOSCA. The dataset includes shapes belonging to different classes (dog, cat, centaur, horse, wolf). Even though our unsupervised network was trained on {\em human} shapes from FAUST, we still get remarkable performance on non-human categories, as we illustrate on the cat. We compare with a supervised correspondence learning technique~\cite{rodola2014dense}, which was trained for each class separately using shapes from TOSCA itself. Additionally, we compare with several axiomatic models. We show our network prediction before (dashed curve) and after (solid) PMF refinement. Note that after refinement, our network achieves practically perfect matching, surpassing the result obtained by initializing PMF with SHOT \cite{vestnerefficient} by a large margin. The rendered visualization is {\em before} refinement.}
\end{center}
\end{figure}
Having an unsupervised loss grants us the ability to train on datasets without given dense correspondences, or even to optimize on individual pairs. Both methods were demonstrated in the previous experiments. In this subsection, however, we would like to pose a different question: what has our network learned by training on a source dataset, and to which extent this knowledge is transferable to a target one. Transferability between training domains is a long-standing research area that has recently re-gained lots of interest, yet it hasen't been explored as much in the shape analysis community. In the scope of this work we focus on transferring from either the synthetic or scanned FAUST shapes to the either (a) human shapes form Dynamic-FAUST, (b) human shapes form SCAPE, and (c) Animal shapes from TOSCA. 

\begin{figure}
\begin{center}
    \input{./dfaust.tikz}
    \begin{overpic}
    [trim=0cm 0cm 0cm 0cm,clip,width=0.4\linewidth]{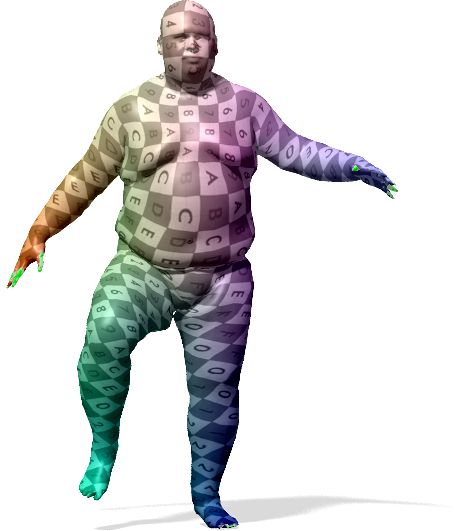}
    \put(20.5,105){\footnotesize Reference}
    \end{overpic}
    \begin{overpic}
    [trim=0cm 0cm 0cm 0cm,clip,width=1.0\linewidth]{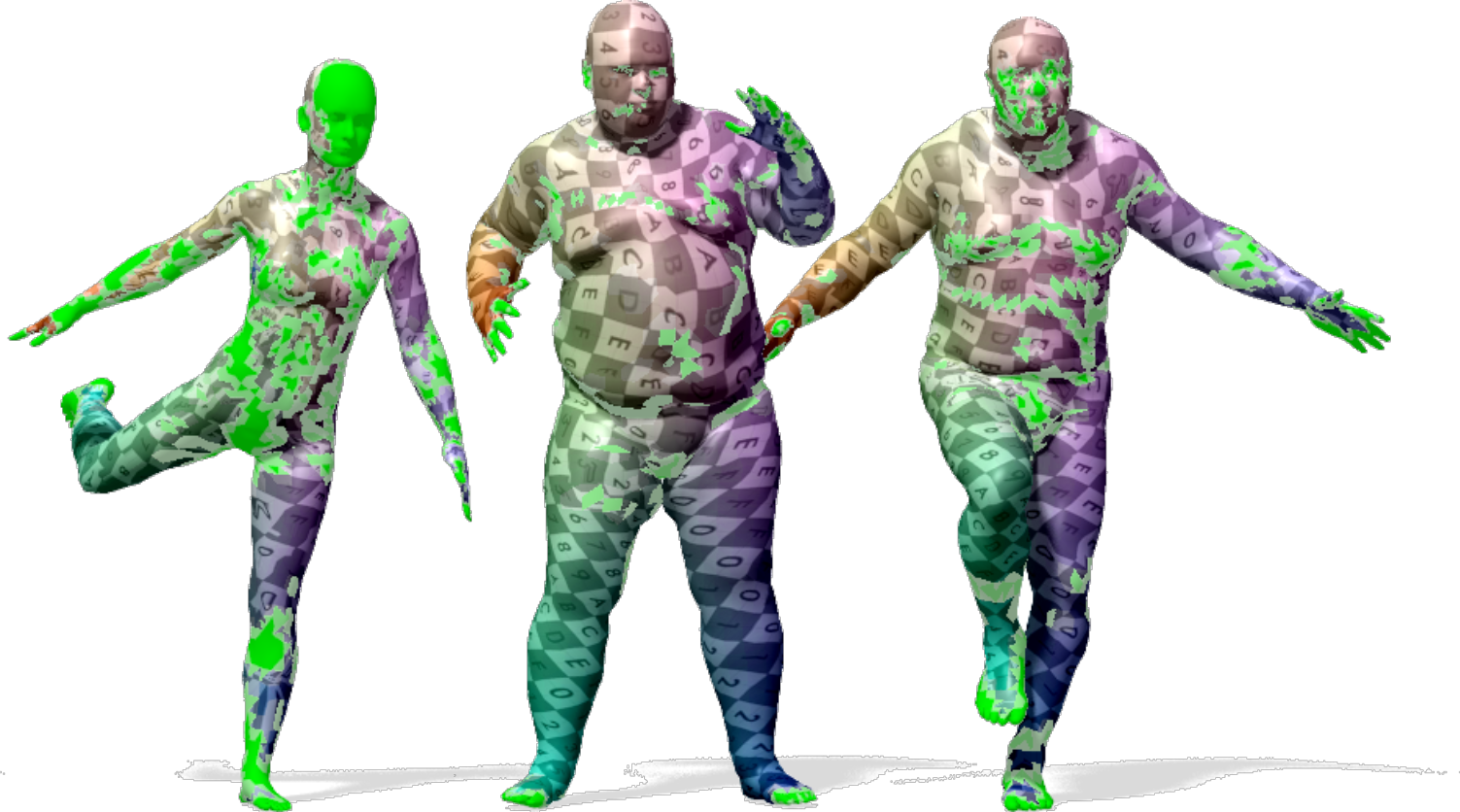}
    \end{overpic}
    \caption{\label{fig:dfaust}Generalization experiments on Dynamic FAUST. We render the {\em raw} network prediction, i.e., without bijectivity-enforcing step; green denotes lack of match, showing that asking for isometry tends to leave the non-isometric parts with no image.}
\end{center}
\end{figure}

\vspace{1ex}\noindent\textbf{Dynamic FAUST} is a recent very large collection of human shapes \cite{dfaust:CVPR:2017}, demonstrating various sequences of activities. While the shapes are triangulated in the same way as our train set of synthetic FAUST, they significantly differ in pose and appearance. Figure \ref{fig:dfaust} shows excellent generalization to this set, suggesting that the small set of $80$ synthetic FAUST shapes were sufficient to capture the pose and shape variability. Please see Appendix \ref{app:DFAUST} for more visualizations.

\vspace{1ex}\noindent\textbf{SCAPE} dataset \cite{scape} also comprises human shapes only. Yet, we've witnessed a quite poor performance using the network trained on synthetic data. By the same reasoning behind the former result, the network might have learned to specialize on the synthetic connectivity. To circumvent this, we have tested the network trained on scans, that demonstrate different meshes. Indeed it can be seen that the generalization improved significantly. 

\vspace{1ex}\noindent\textbf{TOSCA} dataset \cite{bronstein2008numerical} includes various animal shapes. Interestingly, the network trained on scans showed very good performance without ever seeing a single animal shape at train time. In Figure \ref{fig:tosca} we compare the results with a network trained separately on each animal category  and show comparable results before pre-processing, and near-perfect results after.

\begin{figure*}
\begin{center}
    \input{./shrec16.tikz}
    \begin{overpic}
    [trim=0cm 0cm 0cm 0cm,clip,width=0.75\linewidth]{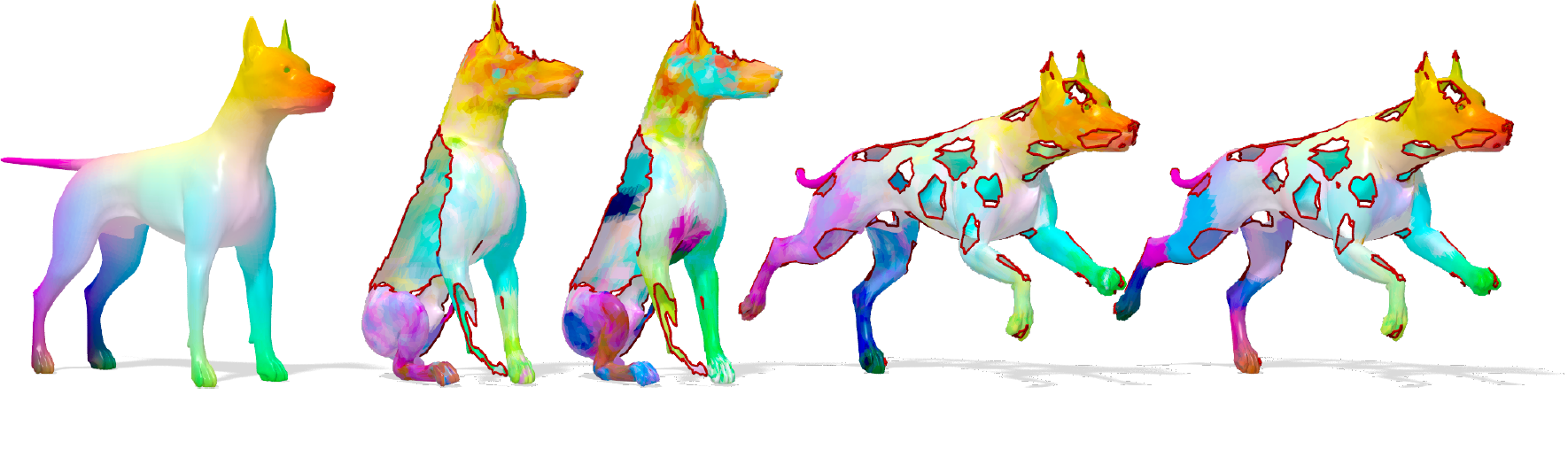}
    \put(8,29.5){\footnotesize Reference}
    \put(29.5,29.5){\footnotesize \textbf{Ours}}
    \put(43,29.5){\footnotesize \cite{rodola2017partial}}
    \put(60,29.5){\footnotesize \textbf{Ours}}
    \put(82,29.5){\footnotesize \cite{rodola2017partial}}
    \end{overpic}
    \caption{\label{fig:partial}Comparisons on the SHREC'16 benchmark~\cite{shrec16-partial} (dog class) for partial matching of deformable shapes. We demonstrate results in line with partial functional maps~\cite{rodola2017partial}, the current state of the art for this problem. The partial shapes shown on the right are matched to the reference; corresponding points have similar color.}
\end{center}
\end{figure*}

\begin{figure}
\begin{center}
    \input{./scape.tikz}
    \caption{\label{fig:scape}Generalization experiments on SCAPE. We show our network prediction without PMF refinement.}
\end{center}
\end{figure}

\subsection{Partial correspondence}
Partial shape correspondence is a notoriously hard problem, and techniques that aim at solving it often require special care \cite{litany2017fully,rodola2017partial,litany2016non}. That said, in this experiment we tested the performance of our method under extreme partiality conditions \textit{as is}, namely, without any modification to our network. To this end, we used the challenging ``dog with holse'' class from \cite{shrec16-partial}. We trained the network on a small set of $10$ partial shapes, and evaluated the results $26$ test shapes. The network results are shown in Figure \ref{fig:partial}.

We found that the mismatches occur typically near the boundary of the partial shape. The reason might be the distortion of the SHOT descriptor in these regions. 

\section{Discussion and conclusions}
The main message of the paper is that a properly designed unsupervised surrogate task can replace massive labeling. 
While we advocate the pure unsupervised approach as a replacement to the supervised one, the two can also be combined in a semi-supervised learning scheme. 
While we demonstrate that the minimization of geodesic distance distortion achieves good generalization on a variety of benchmarks, local scale variation and topological changes can challenge the classic model and require a proper adaptation.
In future studies, we intend to investigate training tasks based on the preservation of more general scale- and conformal-invariant pair-wise geometric quantities, as well as topological properties, e.g. by utilizing pairwise \textit{diffusion} distances.
The proposed network exhibits surprisingly high performance on partial correspondence tasks, even though the functional map layer is not explicitly designed to treat partial data. Extending it to the partial setting based, e.g., on the recently introduced partial functional map formalism \cite{rodola2017partial,litany2017fully} will be the subject of further investigation.
Finally, we would like to explore additional descriptor fields with enhanced properties like increased sensitivity to symmetries, increased robustness to partiality and non rigid deformations. Our work is a first attempt to create a fully unsupervised learning framework to solve the fundamental problem of non rigid shape correspondence. We believe that the fusion of axiomatic models and deep learning is a promising direction that makes it possible to accommodate the expected future growth of 3D data.

\comment{
Unsupervised learning is a proven approach that can calculate very accurate matching in the absence of ground truth labeling and in conditions of a limited size training set.
Future directions, inject different axioms that describes the nature of the correspondence, potentially topological loss term.
\orl{In this paper we focus just on geodesic distances. However we could use other metrics like diffusion or commute time. I wonder if we could learn scale invariant features by training on different sized shapes and using the commute time metric.}
\ron{Or the scale invariant one... Or, all you need to do is modulate your local weight by a Gaussian curvature and you get it. 
In fact, you do not even need to go all the way to fully scale invariant metric. 
There is a simple interpolation between the regular and the scale invariant metric given by $\tidle{g}_{ij} = \kappa^{\alpha} g_{ij}$
 where $\alpha \in [0,1].$
 I would assume that for $\alpha=\sim 0.21$ you'll obtain the best possible  results. 
 BTW for $\alpha =1 $ you have the scale inv. while for $\alpha = 0$ you have the regular metric. 
}

\ron{ In facial surfaces reconstruction from an image, the first to introduce axiomatic models into learning were Richardson et al. \cite{}(3 papers). 
First, by synthesizing rough geometry using the Vetter and Blanz \cite{} 3D morphable model, and then, exploiting the classical shape from shading Lambertian reflectance model as a loss for refinement. Maybe also refer to Tal Hassner and others as a followers.
}
}

\section*{Acknowledgements}
We would like to thank Matteo Sala for providing the models that appear in Figures \ref{fig:teaser} and \ref{fig:hero}. 

This research was partially supported by the Israel Ministry of Science, grant number  3-14719 and the Technion Hiroshi Fujiwara Cyber Security Research Center and the Israel Cyber Bureau. Alex Bronstein was supported by ERC grant no. 335491 (RAPID). Emanuele Rodol\`a was supported by ERC grant no. 802554 (SPECGEO).
\appendix
\begin{appendices}
\input{./appendix.tex}
\end{appendices}

{\small
\bibliographystyle{ieee}
\bibliography{egbib}
}

\end{document}

%% file: tosca.tikz
%
%
\definecolor{mycolor1}{rgb}{0.00000,0.44700,0.74100}%
\definecolor{mycolor2}{rgb}{0.85000,0.32500,0.09800}%
\definecolor{mycolor3}{rgb}{0.92900,0.69400,0.12500}%
\definecolor{mycolor4}{rgb}{0.49400,0.18400,0.55600}%
\definecolor{mycolor5}{rgb}{0.46600,0.67400,0.18800}%
\definecolor{mycolor6}{rgb}{0.30100,0.74500,0.93300}%
\pgfplotsset{scaled y ticks=false}
\pgfplotsset{scaled x ticks=false}
\begin{tikzpicture}

\begin{axis}[%
width=0.44\columnwidth,
height=0.39\columnwidth,
scale only axis,
xmin=0,
xmax=0.1,
xlabel style={font=\color{white!15!black}},
xlabel={\footnotesize Geodesic error},
xlabel style={at={(0.5,0.05)}},
ymin=0,
ymax=100,
ytick={0,20,40,60,80,100},
yticklabels={0,20,40,60,80,100},
xtick={0,0.02,0.04,0.06,0.08,0.1},
xticklabels={0,0.02,0.04,0.06,0.08,0.1},
ylabel style={font=\color{white!15!black}},
ylabel={\footnotesize \% Correspondences},
ylabel style={at={(0.19,0.48)}},
every x tick label/.append style={font=\color{black}, font=\tiny},
every y tick label/.append style={font=\color{black}, font=\tiny},
axis background/.style={fill=white},
axis x line*=bottom,
axis y line*=left,
xmajorgrids,
ymajorgrids,
legend style={at={(0.42,0.05)}, anchor=south west, legend cell align=left, align=left, draw=white!15!black},
title={\footnotesize \textbf{TOSCA}},
title style={at={(0.5,0.96)}}
]
\addplot [color=mycolor1, line width=2.0pt]
  table[row sep=crcr]{%
0	96.8548584820924\\
0.00100000000000477	96.8557644761491\\
0.00199999999999534	96.8589704263928\\
0.00300000000000011	96.8650261101863\\
0.00400000000000489	96.8896050620543\\
0.00499999999999545	96.9280763228893\\
0.00600000000000023	96.9733155588166\\
0.007000000000005	97.0406405139332\\
0.00799999999999557	97.1022660019501\\
0.00900000000000034	97.1571233727859\\
0.0100000000000051	97.2005818094221\\
0.0109999999999957	97.2472461963019\\
0.0120000000000005	97.2864300326133\\
0.0130000000000052	97.3231203520684\\
0.0139999999999958	97.3533986289474\\
0.0150000000000006	97.374415413878\\
0.0160000000000053	97.3897327317088\\
0.0169999999999959	97.4036251827646\\
0.0180000000000007	97.4207235840641\\
0.0190000000000055	97.4363971185886\\
0.019999999999996	97.4545640278805\\
0.0210000000000008	97.4720186458737\\
0.0220000000000056	97.4859110969295\\
0.0229999999999961	97.5051467983914\\
0.0240000000000009	97.5211764075208\\
0.0250000000000057	97.530794116163\\
0.0259999999999962	97.5464675085987\\
0.027000000000001	97.5550167092484\\
0.0280000000000058	97.5721146842814\\
0.0300000000000011	97.5934875438169\\
0.0310000000000059	97.6038176858465\\
0.0319999999999965	97.6187783607171\\
0.0330000000000013	97.6358761936612\\
0.034000000000006	97.6508370106206\\
0.0349999999999966	97.6583174191003\\
0.0360000000000014	97.6725656605834\\
0.0370000000000061	97.6850331027754\\
0.0379999999999967	97.6950066018448\\
0.0390000000000015	97.7096112021105\\
0.0400000000000063	97.7224350030851\\
0.0409999999999968	97.7316964950335\\
0.0420000000000016	97.7484382533727\\
0.0430000000000064	97.7601934042661\\
0.0439999999999969	97.7744419299269\\
0.0450000000000017	97.7840590702137\\
0.0460000000000065	97.7915391945157\\
0.046999999999997	97.8022254111502\\
0.0480000000000018	97.8079247361612\\
0.0490000000000066	97.8161174359396\\
0.0499999999999972	97.8278724447441\\
0.0520000000000067	97.8449704197771\\
0.0529999999999973	97.8524508282568\\
0.054000000000002	97.8585059436951\\
0.0550000000000068	97.8684795848533\\
0.0559999999999974	97.8809468849565\\
0.0570000000000022	97.8887835101299\\
0.0580000000000069	97.8934141850597\\
0.0589999999999975	97.8991135100707\\
0.0600000000000023	97.9076627107204\\
0.061000000000007	97.9140746112077\\
0.0619999999999976	97.9226238118575\\
0.0630000000000024	97.9301040782483\\
0.063999999999993	97.9350909698719\\
0.0649999999999977	97.9422151616578\\
0.0660000000000025	97.9514765115174\\
0.0669999999999931	97.9600254279895\\
0.0679999999999978	97.965012319613\\
0.0690000000000026	97.9685744865504\\
0.0699999999999932	97.973205303569\\
0.070999999999998	97.9810419287424\\
0.0720000000000027	97.987097612536\\
0.0729999999999933	97.9945781631045\\
0.0750000000000028	98.0031272216654\\
0.0759999999999934	98.0059769552153\\
0.0769999999999982	98.0102515555402\\
0.078000000000003	98.0131012890901\\
0.0789999999999935	98.0166634560275\\
0.0799999999999983	98.0223629231273\\
0.0810000000000031	98.0277061735334\\
0.0819999999999936	98.0355429407957\\
0.0829999999999984	98.0412424078955\\
0.0840000000000032	98.0433797080579\\
0.0849999999999937	98.0462294416078\\
0.0859999999999985	98.0515726920139\\
0.0870000000000033	98.0537099921764\\
0.0879999999999939	98.0586970258887\\
0.0889999999999986	98.0643964929885\\
0.0900000000000034	98.0672462265385\\
0.090999999999994	98.0693835267009\\
0.0919999999999987	98.0729456936383\\
0.0930000000000035	98.0757954271882\\
0.0939999999999941	98.0772202939632\\
0.0949999999999989	98.0807824609006\\
0.0969999999999942	98.0907565283252\\
0.097999999999999	98.0921813951002\\
0.0990000000000038	98.0964559954251\\
0.0999999999999943	98.099305728975\\
0.100999999999999	98.1028678959124\\
};
\addlegendentry{\footnotesize \textbf{Ours}}

\addplot [forget plot, color=mycolor1, dashed, line width=2.0pt]
  table[row sep=crcr]{%
0	32.652340646314\\
0.00100000000000477	32.9554599364965\\
0.00199999999999534	33.9599759430372\\
0.00300000000000011	35.7325969160192\\
0.00400000000000489	38.198803743222\\
0.00499999999999545	41.0451974149084\\
0.00600000000000023	43.9834220757481\\
0.007000000000005	46.6897225044083\\
0.00799999999999557	49.2438757178941\\
0.0100000000000051	53.9043790916148\\
0.0109999999999957	56.1319432773418\\
0.0120000000000005	58.2328847073776\\
0.0130000000000052	60.1952386885337\\
0.0139999999999958	62.0315213627856\\
0.0150000000000006	63.7115603621689\\
0.0160000000000053	65.3248838157542\\
0.0169999999999959	66.8389112364239\\
0.0180000000000007	68.2733119826433\\
0.0190000000000055	69.6575423087008\\
0.0210000000000008	72.1821708205011\\
0.0229999999999961	74.4572491034322\\
0.0240000000000009	75.477046371822\\
0.0259999999999962	77.4350485483733\\
0.027000000000001	78.3190299048861\\
0.0280000000000058	79.1769575373389\\
0.0289999999999964	80.0000276319955\\
0.0310000000000059	81.5470805513061\\
0.0330000000000013	83.0406310007659\\
0.0360000000000014	85.0535070304972\\
0.0390000000000015	86.7458598062789\\
0.0420000000000016	88.2615749179336\\
0.0430000000000064	88.7407875927912\\
0.0439999999999969	89.1784048616592\\
0.0460000000000065	89.9873494439768\\
0.046999999999997	90.3514820428854\\
0.0499999999999972	91.3838946518214\\
0.0520000000000067	92.0155545848602\\
0.054000000000002	92.565213863231\\
0.0550000000000068	92.8213936710349\\
0.0559999999999974	93.0617561678594\\
0.0570000000000022	93.2876859214077\\
0.0580000000000069	93.5054550960609\\
0.0600000000000023	93.8848161267788\\
0.061000000000007	94.0482184192139\\
0.0619999999999976	94.2063672482536\\
0.0630000000000024	94.3378449447858\\
0.063999999999993	94.4566246605241\\
0.0649999999999977	94.5663010250073\\
0.0660000000000025	94.6823361636629\\
0.0669999999999931	94.7830989951273\\
0.0679999999999978	94.8718514815935\\
0.0690000000000026	94.951651315255\\
0.0699999999999932	95.0395752690954\\
0.070999999999998	95.1165423145108\\
0.0720000000000027	95.1970748263156\\
0.0729999999999933	95.2707649961739\\
0.0739999999999981	95.3389693260582\\
0.0750000000000028	95.4135615906161\\
0.0759999999999934	95.470652597482\\
0.078000000000003	95.5781752908006\\
0.0789999999999935	95.6277578711128\\
0.0799999999999983	95.6796981091355\\
0.0810000000000031	95.7285977262904\\
0.0819999999999936	95.7799091576548\\
0.0829999999999984	95.8290973627936\\
0.0840000000000032	95.8742254694288\\
0.0849999999999937	95.9083187465743\\
0.0859999999999985	95.9437509392799\\
0.0870000000000033	95.9766424005523\\
0.0879999999999939	96.0127166015851\\
0.0889999999999986	96.0468759465253\\
0.090999999999994	96.1197899546537\\
0.0919999999999987	96.1515945169386\\
0.0930000000000035	96.1868231590491\\
0.0939999999999941	96.2183914606398\\
0.0949999999999989	96.2481212780814\\
0.0960000000000036	96.279134534655\\
0.0969999999999942	96.3023955722703\\
0.097999999999999	96.3296892820786\\
0.0999999999999943	96.3968370785104\\
0.100999999999999	96.4209460553115\\
};

\addplot [color=mycolor2, line width=1.0pt]
  table[row sep=crcr]{%
0	5\\
0.0101156069364094	38.9285714285714\\
0.0303468208092426	83.3866995073892\\
0.0404624277456662	90.4064039408867\\
0.0606936416184993	97.4261083743842\\
0.0708092485549088	97.4261083743842\\
0.0794797687861291	98.5960591133005\\
0.0910404624277419	99.1810344827586\\
0.109826589595372	99.7660098522167\\
};
\addlegendentry{\footnotesize RF~\cite{rodola2014dense}}

\addplot [color=mycolor3,solid,line width=1.0pt]
  table[row sep=crcr]{%
0	43.2027397260274\\
0.001	43.2205479452055\\
0.002	43.4205479452055\\
0.003	44.6698630136986\\
0.004	48.7123287671233\\
0.005	53.7109589041096\\
0.006	58.5808219178082\\
0.007	62.9561643835616\\
0.008	66.7397260273973\\
0.009	70.1397260273973\\
0.01	73.1315068493151\\
0.011	75.6205479452055\\
0.012	77.8643835616438\\
0.013	79.9164383561644\\
0.014	81.6657534246576\\
0.015	83.2520547945206\\
0.016	84.6630136986301\\
0.017	85.9123287671233\\
0.018	87.0493150684931\\
0.019	88.0917808219178\\
0.02	88.9986301369863\\
0.021	89.7479452054795\\
0.022	90.4794520547945\\
0.023	91.1301369863014\\
0.024	91.7232876712329\\
0.025	92.2506849315068\\
0.026	92.6972602739726\\
0.027	93.127397260274\\
0.028	93.5301369863013\\
0.029	93.8780821917808\\
0.03	94.1904109589041\\
0.031	94.5383561643836\\
0.032	94.8219178082191\\
0.033	95.086301369863\\
0.034	95.331506849315\\
0.035	95.5438356164384\\
0.036	95.7493150684932\\
0.037	95.9520547945206\\
0.038	96.1205479452055\\
0.039	96.2794520547945\\
0.04	96.458904109589\\
0.041	96.6191780821918\\
0.042	96.772602739726\\
0.043	96.9027397260274\\
0.044	97.0342465753425\\
0.045	97.1589041095891\\
0.046	97.2739726027397\\
0.047	97.3958904109589\\
0.048	97.5287671232876\\
0.049	97.6260273972602\\
0.05	97.7506849315068\\
0.051	97.827397260274\\
0.052	97.913698630137\\
0.053	97.986301369863\\
0.054	98.058904109589\\
0.055	98.1191780821917\\
0.056	98.1671232876712\\
0.057	98.213698630137\\
0.058	98.2684931506849\\
0.059	98.3041095890411\\
0.06	98.3369863013698\\
0.061	98.3767123287671\\
0.062	98.4109589041095\\
0.063	98.4438356164383\\
0.064	98.4712328767123\\
0.065	98.4876712328767\\
0.066	98.5123287671233\\
0.067	98.5397260273973\\
0.068	98.5561643835616\\
0.069	98.5794520547945\\
0.07	98.5931506849315\\
0.071	98.6041095890411\\
0.072	98.6164383561644\\
0.073	98.6246575342466\\
0.074	98.6342465753425\\
0.075	98.6397260273973\\
0.076	98.6534246575342\\
0.077	98.6671232876713\\
0.078	98.6808219178082\\
0.079	98.6904109589041\\
0.08	98.7\\
0.081	98.7068493150685\\
0.082	98.7164383561644\\
0.083	98.7246575342466\\
0.084	98.7315068493151\\
0.085	98.7465753424657\\
0.086	98.7547945205479\\
0.087	98.7616438356164\\
0.088	98.7753424657534\\
0.089	98.7794520547945\\
0.09	98.7835616438356\\
0.091	98.7876712328767\\
0.092	98.7945205479452\\
0.093	98.8\\
0.094	98.8123287671233\\
0.095	98.8205479452055\\
0.096	98.827397260274\\
0.097	98.8369863013699\\
0.098	98.8424657534247\\
0.099	98.8534246575342\\
0.1	98.8561643835616\\
0.101	98.8657534246575\\
0.102	98.8684931506849\\
0.103	98.8698630136986\\
0.104	98.8767123287671\\
0.105	98.8849315068493\\
0.106	98.8958904109589\\
0.107	98.9082191780822\\
0.108	98.9164383561644\\
0.109	98.9191780821918\\
0.11	98.9205479452055\\
0.111	98.9260273972603\\
0.112	98.9315068493151\\
0.113	98.9328767123288\\
0.114	98.941095890411\\
0.115	98.9465753424657\\
0.116	98.9534246575342\\
0.117	98.958904109589\\
0.118	98.9657534246575\\
0.119	98.9739726027397\\
0.12	98.9835616438356\\
0.121	98.9890410958904\\
0.122	98.9958904109589\\
0.123	99\\
0.124	99.0013698630137\\
0.125	99.0082191780822\\
0.126	99.0164383561644\\
0.127	99.0232876712329\\
0.128	99.027397260274\\
0.129	99.0342465753425\\
0.13	99.0424657534247\\
0.131	99.0479452054795\\
0.132	99.0547945205479\\
0.133	99.0602739726027\\
0.134	99.0616438356164\\
0.135	99.0712328767123\\
0.136	99.072602739726\\
0.137	99.0739726027397\\
0.138	99.0821917808219\\
0.139	99.0849315068493\\
0.14	99.0876712328767\\
0.141	99.0904109589041\\
0.142	99.0945205479452\\
0.143	99.1\\
0.144	99.1013698630137\\
0.145	99.1109589041096\\
0.146	99.113698630137\\
0.147	99.1219178082192\\
0.148	99.1301369863014\\
0.149	99.1356164383562\\
0.15	99.1438356164383\\
0.151	99.1561643835616\\
0.152	99.158904109589\\
0.153	99.1630136986301\\
0.154	99.1712328767123\\
0.155	99.1753424657534\\
0.156	99.1780821917808\\
0.157	99.1849315068493\\
0.158	99.1890410958904\\
0.159	99.1890410958904\\
0.16	99.1945205479452\\
0.161	99.1958904109589\\
0.162	99.2\\
0.163	99.2027397260274\\
0.164	99.2068493150685\\
0.165	99.2123287671233\\
0.166	99.2123287671233\\
0.167	99.2164383561644\\
0.168	99.2219178082192\\
0.169	99.2246575342466\\
0.17	99.2301369863014\\
0.171	99.2369863013699\\
0.172	99.2424657534247\\
0.173	99.2479452054794\\
0.174	99.2534246575342\\
0.175	99.2561643835616\\
0.176	99.2616438356164\\
0.177	99.2643835616438\\
0.178	99.2657534246575\\
0.179	99.2753424657534\\
0.18	99.2808219178082\\
0.181	99.286301369863\\
0.182	99.2904109589041\\
0.183	99.2972602739726\\
0.184	99.3041095890411\\
0.185	99.3095890410959\\
0.186	99.3109589041096\\
0.187	99.313698630137\\
0.188	99.3150684931507\\
0.189	99.3178082191781\\
0.19	99.3191780821918\\
0.191	99.3191780821918\\
0.192	99.3232876712329\\
0.193	99.3301369863014\\
0.194	99.3342465753425\\
0.195	99.3342465753425\\
0.196	99.3369863013699\\
0.197	99.3383561643836\\
0.198	99.3397260273973\\
0.199	99.341095890411\\
0.2	99.3465753424658\\
0.201	99.3465753424658\\
0.202	99.3520547945205\\
0.203	99.3534246575342\\
0.204	99.358904109589\\
0.205	99.3602739726027\\
0.206	99.3616438356164\\
0.207	99.3630136986301\\
0.208	99.3643835616438\\
0.209	99.3657534246575\\
0.21	99.3657534246575\\
0.211	99.3684931506849\\
0.212	99.3684931506849\\
0.213	99.372602739726\\
0.214	99.3767123287671\\
0.215	99.3780821917808\\
0.216	99.3835616438356\\
0.217	99.386301369863\\
0.218	99.3917808219178\\
0.219	99.3931506849315\\
0.22	99.3986301369863\\
0.221	99.4\\
0.222	99.4013698630137\\
0.223	99.4041095890411\\
0.224	99.4054794520548\\
0.225	99.4095890410959\\
0.226	99.4109589041096\\
0.227	99.4123287671233\\
0.228	99.4123287671233\\
0.229	99.413698630137\\
0.23	99.413698630137\\
0.231	99.4164383561644\\
0.232	99.4178082191781\\
0.233	99.4219178082192\\
0.234	99.4219178082192\\
0.235	99.4232876712329\\
0.236	99.4260273972603\\
0.237	99.4287671232877\\
0.238	99.4301369863014\\
0.239	99.4315068493151\\
0.24	99.4328767123288\\
0.241	99.4342465753425\\
0.242	99.4356164383562\\
0.243	99.441095890411\\
0.244	99.4438356164383\\
0.245	99.4493150684931\\
0.246	99.4493150684931\\
0.247	99.4520547945205\\
0.248	99.4547945205479\\
0.249	99.4575342465753\\
0.25	99.4602739726027\\
};
\addlegendentry{\footnotesize{PMF~\cite{vestnerefficient}}}

\addplot [color=mycolor4,solid,line width=1.25pt]
  table[row sep=crcr]{%
0	0.950928\\
0.0025	4.64478\\
0.005	9.49219\\
0.0075	15.0183\\
0.01	21.0095\\
0.0125	27.1948\\
0.015	33.3997\\
0.0175	39.5203\\
0.02	45.3137\\
0.0225	50.6812\\
0.025	55.9033\\
0.0275	60.4321\\
0.03	64.6362\\
0.0325	68.252\\
0.035	71.571\\
0.0375	74.4446\\
0.04	77.002\\
0.0425	79.2944\\
0.045	81.2939\\
0.0475	83.0432\\
0.05	84.6729\\
0.0525	86.2305\\
0.055	87.5488\\
0.0575	88.6487\\
0.06	89.6265\\
0.0625	90.4895\\
0.065	91.4148\\
0.0675	92.146\\
0.07	92.8247\\
0.0725	93.4058\\
0.075	93.9404\\
0.0775	94.3921\\
0.08	94.7803\\
0.0825	95.0964\\
0.085	95.3918\\
0.0875	95.6604\\
0.09	95.8862\\
0.0925	96.084\\
0.095	96.2659\\
0.0975	96.4355\\
0.1	96.593\\
0.1025	96.7273\\
0.105	96.8311\\
0.1075	96.9165\\
0.11	97.0288\\
0.1125	97.1191\\
0.115	97.1826\\
0.1175	97.2656\\
0.12	97.3376\\
0.1225	97.417\\
0.125	97.4915\\
0.1275	97.55\\
0.13	97.6172\\
0.1325	97.6758\\
0.135	97.738\\
0.1375	97.7954\\
0.14	97.8394\\
0.1425	97.8943\\
0.145	97.9468\\
0.1475	97.9944\\
0.15	98.0322\\
0.1525	98.0688\\
0.155	98.1079\\
0.1575	98.158\\
0.16	98.2104\\
0.1625	98.2422\\
0.165	98.291\\
0.1675	98.3374\\
0.17	98.3765\\
0.1725	98.4143\\
0.175	98.4497\\
0.1775	98.4863\\
0.18	98.5205\\
0.1825	98.5706\\
0.185	98.6084\\
0.1875	98.6438\\
0.19	98.6853\\
0.1925	98.7097\\
0.195	98.7549\\
0.1975	98.7817\\
0.2	98.8123\\
0.2025	98.8416\\
0.205	98.8745\\
0.2075	98.8989\\
0.21	98.9197\\
0.2125	98.9417\\
0.215	98.9709\\
0.2175	98.988\\
0.22	99.0088\\
0.2225	99.0417\\
0.225	99.0674\\
0.2275	99.082\\
0.23	99.0979\\
0.2325	99.1138\\
0.235	99.1235\\
0.2375	99.1382\\
0.24	99.1516\\
0.2425	99.1638\\
0.245	99.1797\\
0.2475	99.187\\
};
\addlegendentry{\footnotesize{BIM \cite{kim11}}}

\addplot [color=mycolor5,solid,line width=1.25pt]
  table[row sep=crcr]{%
0	11.5676759834369\\
0.0015	14.3400621118012\\
0.003	20.8993271221532\\
0.0045	27.8416149068323\\
0.006	35.9042874396135\\
0.0075	43.2116977225673\\
0.009	50.7093253968254\\
0.0105	56.4141218081435\\
0.012	61.2409420289855\\
0.0135	64.5876466528641\\
0.015	67.8847481021394\\
0.0165	71.8249654934438\\
0.018	74.1681763285024\\
0.0195	76.8855676328502\\
0.021	78.8860852311939\\
0.0225	80.7153640441684\\
0.024	82.0867839889579\\
0.0255	83.4331866804693\\
0.027	84.1597653554176\\
0.0285	85.3101276742581\\
0.03	85.9845151828847\\
0.0315	86.5821256038647\\
0.033	87.2888630089717\\
0.0345	87.70294168392\\
0.036	88.1344893029676\\
0.0375	88.6824965493444\\
0.039	89.2503450655625\\
0.0405	89.7586697722567\\
0.042	90.1356538992409\\
0.0435	90.3687888198758\\
0.045	90.688621463078\\
0.0465	91.0656055900621\\
0.048	91.367969289165\\
0.0495	91.7669513457557\\
0.051	92.1215062111802\\
0.0525	92.4389665286405\\
0.054	92.9149413388544\\
0.0555	92.9496635610766\\
0.057	93.3266476880607\\
0.0585	93.6389320220842\\
0.06	93.8869478951001\\
0.0615	94.1597653554175\\
0.063	94.3853519668737\\
0.0645	94.514320220842\\
0.066	94.5738440303658\\
0.0675	94.8913043478261\\
0.069	94.9260265700483\\
0.0705	95.0649154589372\\
0.072	95.1938837129055\\
0.0735	95.3228519668737\\
0.075	95.3922964113182\\
0.0765	95.4889147688061\\
0.078	95.5236369910283\\
0.0795	95.7145013802623\\
0.081	96.1262077294687\\
0.0825	96.443668046929\\
0.084	96.4783902691512\\
0.0855	96.632160110421\\
0.087	96.8206521739131\\
0.0885	96.8900966183575\\
0.09	97.0091442374051\\
0.0915	97.162914078675\\
0.093	97.2323585231194\\
0.0945	97.2670807453416\\
0.096	97.3266045548654\\
0.0975	97.3861283643893\\
0.099	97.3861283643893\\
0.1005	97.4803743961353\\
0.102	97.4803743961353\\
0.1035	97.5150966183575\\
0.105	97.5150966183575\\
0.1065	97.5150966183575\\
0.108	97.5150966183575\\
0.1095	97.6117149758454\\
0.111	97.6464371980676\\
0.1125	97.6464371980676\\
0.114	97.6464371980676\\
0.1155	97.7406832298137\\
0.117	97.7406832298137\\
0.1185	97.7406832298137\\
0.12	97.8002070393375\\
0.1215	97.8002070393375\\
0.123	97.9192546583851\\
0.1245	97.9539768806073\\
0.126	97.9539768806073\\
0.1275	97.9539768806073\\
0.129	98.0135006901311\\
0.1305	98.0135006901311\\
0.132	98.0135006901311\\
0.1335	98.1077467218772\\
0.135	98.1077467218772\\
0.1365	98.1424689440994\\
0.138	98.2367149758454\\
0.1395	98.2367149758454\\
0.141	98.2367149758454\\
0.1425	98.3309610075915\\
0.144	98.4252070393375\\
0.1455	98.4599292615597\\
0.147	98.4599292615597\\
0.1485	98.4946514837819\\
0.15	98.4946514837819\\
0.1515	98.4946514837819\\
0.153	98.5293737060042\\
0.1545	98.588897515528\\
0.156	98.588897515528\\
0.1575	98.588897515528\\
0.159	98.6484213250518\\
0.1605	98.6484213250518\\
0.162	98.6484213250518\\
0.1635	98.683143547274\\
0.165	98.683143547274\\
0.1665	98.683143547274\\
0.168	98.7178657694962\\
0.1695	98.7178657694962\\
0.171	98.7525879917184\\
0.1725	98.7873102139407\\
0.174	98.7873102139407\\
0.1755	98.7873102139407\\
0.177	98.7873102139407\\
0.1785	98.7873102139407\\
0.18	98.7873102139407\\
0.1815	98.7873102139407\\
0.183	98.7873102139407\\
0.1845	98.9063578329883\\
0.186	98.9063578329883\\
0.1875	98.9658816425121\\
0.189	98.9658816425121\\
0.1905	98.9658816425121\\
0.192	98.9658816425121\\
0.1935	98.9658816425121\\
0.195	99.0254054520359\\
0.1965	99.0601276742581\\
0.198	99.0601276742581\\
0.1995	99.0601276742581\\
0.201	99.0601276742581\\
0.2025	99.0601276742581\\
0.204	99.1196514837819\\
0.2055	99.1196514837819\\
0.207	99.1543737060042\\
0.2085	99.1543737060042\\
0.21	99.1543737060042\\
0.2115	99.1543737060042\\
0.213	99.1543737060042\\
0.2145	99.1543737060042\\
0.216	99.1543737060042\\
0.2175	99.1543737060042\\
0.219	99.1543737060042\\
0.2205	99.1543737060042\\
0.222	99.1543737060042\\
0.2235	99.1890959282264\\
0.225	99.1890959282264\\
0.2265	99.1890959282264\\
0.228	99.1890959282264\\
0.2295	99.1890959282264\\
0.231	99.1890959282264\\
0.2325	99.1890959282264\\
0.234	99.1890959282264\\
0.2355	99.1890959282264\\
0.237	99.2486197377502\\
0.2385	99.2486197377502\\
0.24	99.2486197377502\\
0.2415	99.2486197377502\\
0.243	99.2486197377502\\
0.2445	99.2486197377502\\
0.246	99.2486197377502\\
0.2475	99.308143547274\\
0.249	99.308143547274\\
};
\addlegendentry{\footnotesize{SGMDS \cite{aflalo2016spectral}}}

\addplot [color=mycolor6,solid,line width=1.25pt]
  table[row sep=crcr]{%
0	21.0526315789474\\
0.00114810562571757	22.9018492176387\\
0.00172215843857635	24.75106685633\\
0.00229621125143513	26.6002844950213\\
0.00287026406429392	28.5917496443812\\
0.00315729047072331	30.298719772404\\
0.00373134328358209	32.2901849217639\\
0.00430539609644087	34.1394025604552\\
0.00487944890929966	36.1308677098151\\
0.00545350172215844	37.8378378378378\\
0.00602755453501722	39.8293029871977\\
0.00660160734787601	41.5362731152205\\
0.00717566016073479	43.5277382645804\\
0.00832376578645235	45.3769559032717\\
0.00889781859931114	46.9416785206259\\
0.0100459242250287	48.7908961593172\\
0.0106199770378875	50.49786628734\\
0.0111940298507463	51.778093883357\\
0.0120551090700344	53.2005689900427\\
0.0126291618828932	54.7652916073969\\
0.0134902411021814	56.0455192034139\\
0.0143513203214696	57.7524893314367\\
0.0149253731343284	59.3172119487909\\
0.0160734787600459	61.0241820768137\\
0.0172215843857635	62.7311522048364\\
0.0183696900114811	64.2958748221906\\
0.0195177956371986	65.8605974395448\\
0.0206659012629162	67.2830725462304\\
0.0215269804822044	68.8477951635846\\
0.0229621125143513	70.2702702702703\\
0.0241102181400689	71.4082503556188\\
0.0255453501722158	72.972972972973\\
0.0269804822043628	74.25320056899\\
0.0287026406429392	75.3911806543386\\
0.0301377726750861	76.5291607396871\\
0.0315729047072331	77.6671408250356\\
0.0335820895522388	78.9473684210526\\
0.0350172215843858	80.0853485064011\\
0.0373134328358209	81.3655761024182\\
0.0390355912743972	82.5035561877667\\
0.041044776119403	83.4992887624467\\
0.0433409873708381	84.7795163584637\\
0.0456371986222733	85.7752489331437\\
0.0482204362801378	86.7709815078236\\
0.0510907003444317	87.7667140825036\\
0.054247990815155	88.4779516358464\\
0.0574052812858783	89.3314366998578\\
0.0605625717566016	90.3271692745377\\
0.0637198622273249	90.7539118065434\\
0.067451205510907	91.3229018492176\\
0.0714695752009185	92.0341394025605\\
0.0749138920780712	92.0341394025605\\
0.0783582089552239	92.4608819345662\\
0.0812284730195178	93.0298719772404\\
0.0849598163030999	93.4566145092461\\
0.0889781859931114	93.7411095305832\\
0.0932835820895522	94.0256045519203\\
0.0967278989667049	94.3100995732575\\
0.100746268656716	94.452347083926\\
0.104477611940299	94.5945945945946\\
0.107634902411022	94.8790896159317\\
0.111653272101033	95.3058321479374\\
0.115384615384615	95.5903271692745\\
0.119402985074627	95.8748221906117\\
0.12284730195178	96.0170697012802\\
0.12743972445465	96.3015647226173\\
0.131458094144661	96.5860597439545\\
0.134615384615385	96.5860597439545\\
0.138059701492537	96.7283072546231\\
0.140929965556831	96.5860597439545\\
0.144087256027555	96.7283072546231\\
0.147531572904707	96.7283072546231\\
0.151549942594719	96.7283072546231\\
0.154994259471871	96.8705547652916\\
0.159299655568312	97.0128022759602\\
0.163030998851894	97.0128022759602\\
0.167336394948335	96.8705547652916\\
0.171354764638347	96.8705547652916\\
0.175373134328358	97.0128022759602\\
0.179678530424799	97.0128022759602\\
0.18398392652124	97.1550497866287\\
0.188289322617681	97.1550497866287\\
0.192020665901263	97.1550497866287\\
0.196900114810563	97.2972972972973\\
0.200918484500574	97.2972972972973\\
0.205223880597015	97.4395448079659\\
0.209242250287026	97.4395448079659\\
0.213260619977038	97.2972972972973\\
0.21699196326062	97.4395448079659\\
0.221297359357061	97.4395448079659\\
0.224741676234214	97.5817923186344\\
0.228473019517796	97.5817923186344\\
0.232204362801378	97.5817923186344\\
0.23593570608496	97.724039829303\\
0.239380022962113	97.724039829303\\
0.243398392652124	97.8662873399716\\
0.247416762342135	97.724039829303\\
0.25	97.8662873399716\\
};
\addlegendentry{\footnotesize{FM \cite{ovsjanikov12}}}

\end{axis}
\end{tikzpicture}%

%% file: shrec16.tikz
%
%
\definecolor{mycolor1}{rgb}{0.00000,0.44700,0.74100}%
\definecolor{mycolor2}{rgb}{0.85000,0.32500,0.09800}%
\pgfplotsset{scaled y ticks=false}
\pgfplotsset{scaled x ticks=false}
\begin{tikzpicture}

\begin{axis}[%
width=0.38\columnwidth,
height=0.39\columnwidth,
scale only axis,
xmin=0,
xmax=0.1,
xlabel style={font=\color{white!15!black}},
xlabel={\footnotesize Geodesic error},
xlabel style={at={(0.5,0.05)}},
ytick={0,20,40,60,80,100},
yticklabels={0,20,40,60,80,100},
xtick={0,0.02,0.04,0.06,0.08,0.1},
xticklabels={0,0.02,0.04,0.06,0.08,0.1},
ymin=0,
ymax=100,
ylabel style={font=\color{white!15!black}},
ylabel={\footnotesize \% Correspondences},
ylabel style={at={(0.19,0.48)}},
every x tick label/.append style={font=\color{black}, font=\tiny},
every y tick label/.append style={font=\color{black}, font=\tiny},
axis background/.style={fill=white},
axis x line*=bottom,
axis y line*=left,
xmajorgrids,
ymajorgrids,
legend style={at={(0.11,0.05)}, anchor=south west, legend cell align=left, align=left, draw=white!15!black}
]

\addplot [color=mycolor2, line width=2.0pt]
  table[row sep=crcr]{%
0	2.87215817163339\\
0.00100000000000477	2.87390394404713\\
0.00199999999999534	2.94826539939028\\
0.00300000000000011	3.23690181455623\\
0.00400000000000489	4.02894465854666\\
0.00499999999999545	5.17612168189416\\
0.00600000000000023	6.74309551044335\\
0.007000000000005	8.65103881967113\\
0.00799999999999557	10.9894500652296\\
0.0100000000000051	16.0029165956004\\
0.0109999999999957	18.7404789310996\\
0.0120000000000005	21.3867619776135\\
0.0130000000000052	23.8002874019457\\
0.0160000000000053	30.777473120927\\
0.0169999999999959	32.9775340614285\\
0.0180000000000007	35.111379033737\\
0.0210000000000008	41.2077583873498\\
0.0220000000000056	43.0726382680763\\
0.0240000000000009	46.6010747583039\\
0.0250000000000057	48.1882434067706\\
0.0259999999999962	49.6711846501586\\
0.027000000000001	51.0851278763669\\
0.0280000000000058	52.4456539490526\\
0.0289999999999964	53.7252019758198\\
0.0310000000000059	56.0026487790334\\
0.0319999999999965	57.0900267038333\\
0.0330000000000013	58.13180227945\\
0.0349999999999966	59.9966657238485\\
0.0370000000000061	61.8057283314454\\
0.0379999999999967	62.6621075751113\\
0.0400000000000063	64.2063273343926\\
0.0420000000000016	65.5937687330846\\
0.0430000000000064	66.2593977140678\\
0.0439999999999969	66.8991253749361\\
0.0460000000000065	68.0762408700078\\
0.046999999999997	68.6632769466365\\
0.0490000000000066	69.7199730505317\\
0.0510000000000019	70.6565681089207\\
0.0520000000000067	71.1153923656458\\
0.0529999999999973	71.5033055970188\\
0.054000000000002	71.8590937484855\\
0.0550000000000068	72.2679087432819\\
0.0559999999999974	72.6254561150918\\
0.0570000000000022	72.9358701095988\\
0.0589999999999975	73.5910286326609\\
0.0600000000000023	73.8888341728314\\
0.061000000000007	74.1954872990331\\
0.0619999999999976	74.4640408789118\\
0.063999999999993	74.9747969946554\\
0.0660000000000025	75.4454076849376\\
0.0669999999999931	75.6808475792594\\
0.0679999999999978	75.9402821390533\\
0.0690000000000026	76.1485028625104\\
0.0699999999999932	76.3361151939867\\
0.070999999999998	76.517214266154\\
0.0720000000000027	76.6898212214052\\
0.0729999999999933	76.8293886000467\\
0.0739999999999981	76.9840552931226\\
0.0750000000000028	77.1019662655059\\
0.0759999999999934	77.2324980466281\\
0.0769999999999982	77.3578014495389\\
0.078000000000003	77.4696116977984\\
0.0789999999999935	77.5575801496845\\
0.0810000000000031	77.7572345916734\\
0.0819999999999936	77.8425506100478\\
0.0840000000000032	78.0220795659196\\
0.0849999999999937	78.0954919541324\\
0.0859999999999985	78.196375917168\\
0.0870000000000033	78.2621153643028\\
0.0879999999999939	78.3325932112444\\
0.0900000000000034	78.4818532079583\\
0.090999999999994	78.5326598903214\\
0.0919999999999987	78.6020287390309\\
0.0930000000000035	78.6585082057863\\
0.0939999999999941	78.7020462583687\\
0.0949999999999989	78.7765251970177\\
0.0960000000000036	78.8304348150875\\
0.0969999999999942	78.8878352427508\\
0.097999999999999	78.938452104332\\
0.0990000000000038	78.9838856687374\\
0.0999999999999943	79.0379448047137\\
0.100999999999999	79.083890814434\\
};
\addlegendentry{\footnotesize \textbf{Ours}}

\addplot [color=mycolor1, line width=1.0pt]
  table[row sep=crcr]{%
0	2.57146390269507\\
0.000900000000001455	2.57146390269507\\
0.00100000000000477	2.57321160994591\\
0.00109999999999388	2.57408469454714\\
0.0011999999999972	2.57992524471069\\
0.00130000000000052	2.58183913949014\\
0.00140000000000384	2.59572856954576\\
0.00149999999999295	2.5994137397837\\
0.00159999999999627	2.61903599801734\\
0.00180000000000291	2.67814742117899\\
0.0023000000000053	2.83375131481125\\
0.00249999999999773	2.9425881792132\\
0.00270000000000437	3.03304763838352\\
0.00300000000000011	3.36983153444093\\
0.00310000000000343	3.47691257499633\\
0.00320000000000675	3.63739046600526\\
0.00339999999999918	3.83968770440133\\
0.00430000000000064	5.21934086327975\\
0.00459999999999638	5.83561814853029\\
0.00490000000000634	6.38142420448726\\
0.00530000000000541	7.19112145927286\\
0.00589999999999691	8.54349635244188\\
0.00660000000000593	10.0569405289815\\
0.00669999999999504	10.3282525304752\\
0.00690000000000168	10.7397084794778\\
0.00790000000000646	13.5137081211995\\
0.00799999999999557	13.7527665284964\\
0.00830000000000553	14.622100532237\\
0.00889999999999702	16.2853395645007\\
0.00920000000000698	17.0154166579677\\
0.00950000000000273	17.9298607607385\\
0.00960000000000605	18.1646885850346\\
0.011099999999999	22.637487367804\\
0.0114999999999981	23.6530014075763\\
0.0121000000000038	25.3213385975287\\
0.0130999999999943	27.6963606545167\\
0.0151000000000039	32.3357182206175\\
0.0152999999999963	32.8089853630065\\
0.0164999999999935	35.4602590563756\\
0.0169000000000068	36.2460593288682\\
0.0172000000000025	36.8129661822636\\
0.0178999999999974	38.1713982687349\\
0.0182999999999964	38.9154770568175\\
0.0186000000000064	39.5075420983807\\
0.0187999999999988	39.8390173018355\\
0.0190000000000055	40.2367569337413\\
0.0215000000000032	44.7962522491976\\
0.0217999999999989	45.3101176888994\\
0.022199999999998	46.0366304606429\\
0.0230999999999995	47.4881673380237\\
0.0233000000000061	47.8163711341547\\
0.0237000000000052	48.3981881068791\\
0.0238999999999976	48.7059803607849\\
0.0241000000000042	49.0081839494775\\
0.0246000000000066	49.6980020679955\\
0.0250999999999948	50.4543902967761\\
0.0254999999999939	51.0195041481209\\
0.0258999999999929	51.5558441867147\\
0.0262000000000029	51.9327233898549\\
0.0263999999999953	52.1649960013834\\
0.0267999999999944	52.7405622637254\\
0.027000000000001	52.9855114724882\\
0.0274000000000001	53.503695608694\\
0.0275000000000034	53.5960234381175\\
0.0276999999999958	53.8614364790978\\
0.0280000000000058	54.2286116184621\\
0.0284000000000049	54.7489412474783\\
0.0285999999999973	54.9405665098989\\
0.0290999999999997	55.5333717566874\\
0.029200000000003	55.6331276858421\\
0.0293999999999954	55.9011808797011\\
0.0296000000000021	56.119980872317\\
0.0297999999999945	56.3371022168357\\
0.030699999999996	57.2982002734382\\
0.0307999999999993	57.3846181355107\\
0.0310000000000059	57.6031372542391\\
0.031099999999995	57.6771418000902\\
0.0313000000000017	57.9004478122599\\
0.0314999999999941	58.0833713072241\\
0.0327000000000055	59.3301817260915\\
0.0328999999999979	59.4961567254637\\
0.0348000000000042	61.1557161151141\\
0.0349999999999966	61.3131267116981\\
0.0352000000000032	61.4712834660581\\
0.0353000000000065	61.5362213256267\\
0.035899999999998	62.0642476937403\\
0.0361000000000047	62.2480760124789\\
0.0362999999999971	62.4312283480906\\
0.0374999999999943	63.4021719046177\\
0.037700000000001	63.5716938669501\\
0.0378999999999934	63.7438636677621\\
0.0384999999999991	64.1716755765122\\
0.0391000000000048	64.5818428724265\\
0.0391999999999939	64.6382230533196\\
0.0396999999999963	65.0147826756766\\
0.0399000000000029	65.16644774292\\
0.0400999999999954	65.3167743194938\\
0.0404000000000053	65.4710618067279\\
0.0407000000000011	65.6613368345115\\
0.0411000000000001	65.8889494074831\\
0.0418999999999983	66.4526114666208\\
0.042199999999994	66.6693303463524\\
0.0427999999999997	67.0578986733871\\
0.0431999999999988	67.2883422994531\\
0.0434999999999945	67.4506037226249\\
0.0437000000000012	67.5923673703729\\
0.0442000000000036	67.859781947403\\
0.0452999999999975	68.5839667282228\\
0.0464000000000055	69.2249480731233\\
0.0467000000000013	69.3756119513551\\
0.046999999999997	69.5416277884718\\
0.0476000000000028	69.8683323330451\\
0.0481000000000051	70.0691450535238\\
0.0481999999999942	70.1085244287065\\
0.0485999999999933	70.3297635841732\\
0.0486999999999966	70.3660669672143\\
0.0489000000000033	70.4731177288356\\
0.0490999999999957	70.5931985290157\\
0.0493000000000023	70.6873267459194\\
0.0497000000000014	70.8721328420647\\
0.0498000000000047	70.9334689736309\\
0.0502000000000038	71.1038969214468\\
0.0511000000000053	71.5926434229473\\
0.051400000000001	71.7350562340731\\
0.0515999999999934	71.8523815036925\\
0.0519000000000034	72.0111128514594\\
0.0523000000000025	72.1872261758882\\
0.0529999999999973	72.4654110047571\\
0.0532000000000039	72.5408775941996\\
0.0534999999999997	72.6512672564961\\
0.0538999999999987	72.8425852875305\\
0.0544000000000011	73.0421711085057\\
0.0545999999999935	73.1172638789743\\
0.0546999999999969	73.1792586438176\\
0.0549000000000035	73.2505105556781\\
0.0550999999999959	73.3415748745662\\
0.0554000000000059	73.4608598967044\\
0.0555999999999983	73.5520344122202\\
0.055800000000005	73.6088982789793\\
0.0559999999999974	73.6941261151686\\
0.0561000000000007	73.7341986626244\\
0.056200000000004	73.7568293282418\\
0.0567000000000064	73.9493068881235\\
0.0570000000000022	74.0531839805959\\
0.0579000000000036	74.388516528729\\
0.058099999999996	74.4606513838392\\
0.0584999999999951	74.5973439322558\\
0.0588999999999942	74.7153600672572\\
0.0591000000000008	74.7946621261911\\
0.0594999999999999	74.9137626356442\\
0.0596000000000032	74.9651501380092\\
0.0601999999999947	75.1538181879284\\
0.0605000000000047	75.2463357566693\\
0.0608000000000004	75.3434499579595\\
0.0613000000000028	75.4991014833784\\
0.0614000000000061	75.5232792408753\\
0.0614999999999952	75.5627663424663\\
0.0621000000000009	75.7300253747206\\
0.0623999999999967	75.8355882412523\\
0.0626000000000033	75.8887790283783\\
0.0627000000000066	75.909379664243\\
0.0627999999999957	75.950944006241\\
0.0644000000000062	76.4113435596749\\
0.0645999999999987	76.4468596217246\\
0.064700000000002	76.4863188889153\\
0.0661000000000058	76.8701558849572\\
0.0662999999999982	76.9070166031259\\
0.0665000000000049	76.9445344423006\\
0.066900000000004	77.0440893360154\\
0.0671999999999997	77.1209254994511\\
0.0675999999999988	77.2407720295817\\
0.0677000000000021	77.2583272532741\\
0.0687000000000069	77.4905782265484\\
0.0688999999999993	77.5471134256717\\
0.0691000000000059	77.5881400593512\\
0.0692999999999984	77.6304808507159\\
0.0694000000000017	77.6467039750575\\
0.0696999999999974	77.7098457487517\\
0.0698000000000008	77.7290611688734\\
0.0699000000000041	77.7581803535667\\
0.0701999999999998	77.8127926363453\\
0.0703000000000031	77.842571701789\\
0.0705999999999989	77.8919963629072\\
0.0712000000000046	78.0234786730463\\
0.071399999999997	78.0549549315725\\
0.071700000000007	78.1139823892892\\
0.0717999999999961	78.128057191326\\
0.0721999999999952	78.2145167264398\\
0.0728000000000009	78.2962579072554\\
0.0729999999999933	78.3225932844512\\
0.0730999999999966	78.3309715884775\\
0.0731999999999999	78.3458429874326\\
0.0733000000000033	78.3656798531503\\
0.0734000000000066	78.3797334824043\\
0.0734999999999957	78.3876724018871\\
0.073599999999999	78.4073710787795\\
0.0737000000000023	78.4179394861318\\
0.0738000000000056	78.4243969984801\\
0.0741000000000014	78.4811230508542\\
0.0742999999999938	78.5009424216617\\
0.0746000000000038	78.5515460453988\\
0.0747000000000071	78.5603751348554\\
0.0748999999999995	78.5931258005498\\
0.0750000000000028	78.6126309684773\\
0.0752999999999986	78.6487393943165\\
0.0755999999999943	78.6863460496933\\
0.0756999999999977	78.6984387873289\\
0.075800000000001	78.7064488650396\\
0.0759999999999934	78.7362803686181\\
0.0760999999999967	78.7438712551673\\
0.0764999999999958	78.7956200439432\\
0.0771000000000015	78.8889621961634\\
0.0772000000000048	78.8971534147105\\
0.0776000000000039	78.9460801032194\\
0.077699999999993	78.962479869579\\
0.0777999999999963	78.9716442570252\\
0.078000000000003	78.9953720998123\\
0.0781000000000063	79.0026827989845\\
0.078400000000002	79.0414297609252\\
0.0785000000000053	79.051536022432\\
0.0785999999999945	79.0579262407481\\
0.0786999999999978	79.0698931257343\\
0.0788000000000011	79.095213630732\\
0.0789999999999935	79.1104161333473\\
0.0793000000000035	79.1383806862256\\
0.0795999999999992	79.1721983504733\\
0.079899999999995	79.2123860535662\\
0.0801000000000016	79.2413520587322\\
0.0802000000000049	79.2491899566489\\
0.080299999999994	79.2632125507753\\
0.0805000000000007	79.2793751690968\\
0.0807999999999964	79.3046780500697\\
0.0811000000000064	79.3478991450051\\
0.0814000000000021	79.3709445287392\\
0.0816999999999979	79.4000870077134\\
0.0819000000000045	79.4177085414525\\
0.0819999999999936	79.4301671867987\\
0.082099999999997	79.4343360316351\\
0.0822000000000003	79.4504622738847\\
0.0823000000000036	79.4575765277359\\
0.0824000000000069	79.4703295639704\\
0.082499999999996	79.4783299208913\\
0.0827000000000027	79.5070848209642\\
0.0828999999999951	79.5255341291201\\
0.0829999999999984	79.5418996828167\\
0.0831000000000017	79.5532344854716\\
0.0832999999999942	79.5691841825627\\
0.0838999999999999	79.6435526993074\\
0.0840000000000032	79.6647194257592\\
0.0841000000000065	79.6719030611445\\
0.0844000000000023	79.7114132084447\\
0.0845000000000056	79.7192499462477\\
0.084699999999998	79.7375308360571\\
0.0848000000000013	79.7450863073105\\
0.0849999999999937	79.7716586587789\\
0.0850999999999971	79.7784917879476\\
0.0852000000000004	79.8008950079651\\
0.085400000000007	79.8178575726901\\
0.0854999999999961	79.8257562030821\\
0.0855999999999995	79.8304575657622\\
0.0857000000000028	79.8448952026402\\
0.0858000000000061	79.8531400938361\\
0.0862000000000052	79.8992950675899\\
0.0865000000000009	79.9310267192053\\
0.0866000000000042	79.9512164990531\\
0.0871000000000066	79.993685541635\\
0.0874000000000024	80.0226503300638\\
0.0875000000000057	80.0309515027883\\
0.0875999999999948	80.046817419418\\
0.0876999999999981	80.0523906795598\\
0.0878000000000014	80.0627858540645\\
0.0883000000000038	80.1002209167312\\
0.0884999999999962	80.1252235074376\\
0.0888999999999953	80.1612379912931\\
0.0889999999999986	80.1704611422676\\
0.089100000000002	80.1755576547763\\
0.0899000000000001	80.2437817735793\\
0.0900000000000034	80.2476778109939\\
0.0901999999999958	80.2694713480018\\
0.0904000000000025	80.2881414404548\\
0.0905000000000058	80.3003721529248\\
0.0905999999999949	80.3161431296303\\
0.0906999999999982	80.3209436089429\\
0.090999999999994	80.3580368124346\\
0.0912000000000006	80.373333440953\\
0.0913000000000039	80.38766319981\\
0.0915999999999997	80.4126992207695\\
0.091700000000003	80.4301823868058\\
0.0918999999999954	80.4489717722049\\
0.0925000000000011	80.521324989902\\
0.0926999999999936	80.5393884443713\\
0.0929000000000002	80.5548690486105\\
0.0930000000000035	80.563843289782\\
0.0931000000000068	80.5781238115711\\
0.0932999999999993	80.5920034217253\\
0.0934000000000026	80.5963282893152\\
0.0935000000000059	80.6026118625328\\
0.093599999999995	80.6139300181313\\
0.093900000000005	80.655442858438\\
0.0939999999999941	80.6586878630951\\
0.0944999999999965	80.7024062737217\\
0.0945999999999998	80.7164139002271\\
0.0947000000000031	80.7227483245967\\
0.0948000000000064	80.7315413945863\\
0.0948999999999955	80.7470264502762\\
0.0951000000000022	80.7637444579186\\
0.0952000000000055	80.7779123509234\\
0.0953999999999979	80.7933533481152\\
0.0956000000000046	80.8208392836405\\
0.0956999999999937	80.8407449850855\\
0.0959000000000003	80.8557657210798\\
0.096100000000007	80.8785556605648\\
0.0961999999999961	80.8865047344642\\
0.0962999999999994	80.8902940132587\\
0.0965999999999951	80.9179897167965\\
0.0966999999999985	80.9220834355875\\
0.0968000000000018	80.9379506932414\\
0.0969999999999942	80.9574215923737\\
0.0973000000000042	80.9856277341507\\
0.0973999999999933	80.9994648260864\\
0.0975999999999999	81.0126048639174\\
0.0977000000000032	81.0184697010253\\
0.0978000000000065	81.0274638358794\\
0.0982999999999947	81.0564585493269\\
0.0986000000000047	81.0950962359799\\
0.0989000000000004	81.1155083354926\\
0.0990000000000038	81.1277071093895\\
0.0992999999999995	81.1546676300743\\
0.0994000000000028	81.1662883482068\\
0.0995999999999952	81.1792621535535\\
0.0996999999999986	81.1926474447786\\
0.0998000000000019	81.1994968564475\\
0.0999999999999943	81.2213582620871\\
};
\addlegendentry{\footnotesize Partial FM~\cite{rodola2017partial}}

\end{axis}
\end{tikzpicture}%

%% file: dfaust.tikz
%
%
\definecolor{mycolor1}{rgb}{0.00000,0.44700,0.74100}%
\begin{tikzpicture}

\begin{axis}[%
width=0.44\columnwidth,
height=0.39\columnwidth,
scale only axis,
xmin=0,
xmax=0.1,
xlabel style={font=\color{white!15!black}},
xlabel={\footnotesize Geodesic error},
xlabel style={at={(0.5,0.05)}},
ymin=0,
ymax=100,
ytick={0,20,40,60,80,100},
yticklabels={0,20,40,60,80,100},
xtick={0,0.02,0.04,0.06,0.08,0.1},
xticklabels={0,0.02,0.04,0.06,0.08,0.1},
ylabel style={font=\color{white!15!black}},
ylabel={\footnotesize \% Correspondences},
ylabel style={at={(0.19,0.48)}},
every x tick label/.append style={font=\color{black}, font=\tiny},
every y tick label/.append style={font=\color{black}, font=\tiny},
axis background/.style={fill=white},
axis x line*=bottom,
axis y line*=left,
xmajorgrids,
ymajorgrids,
legend style={at={(0.46,0.05)}, anchor=south west, legend cell align=left, align=left, draw=white!15!black},
title={\footnotesize \textbf{Dynamic FAUST}},
title style={at={(0.5,0.96)}}
]
\addplot [color=mycolor1, line width=2.0pt]
  table[row sep=crcr]{%
0	48.2640148766328\\
0.00100000000000477	48.5660830914369\\
0.00199999999999534	49.9770954281567\\
0.00300000000000011	52.2498639332366\\
0.00400000000000489	54.8244738751814\\
0.00499999999999545	57.4761883164006\\
0.00600000000000023	60.3276941219158\\
0.007000000000005	63.4833091436865\\
0.00900000000000034	69.9417180696662\\
0.0100000000000051	73.0317942670537\\
0.0109999999999957	75.770137880987\\
0.0120000000000005	78.2116291727141\\
0.0130000000000052	80.4723784470247\\
0.0139999999999958	82.4970518867924\\
0.0150000000000006	84.3235214078375\\
0.0160000000000053	85.9449836719884\\
0.0169999999999959	87.3199383164006\\
0.0180000000000007	88.5524764150943\\
0.0190000000000055	89.6507619738752\\
0.019999999999996	90.603456095791\\
0.0210000000000008	91.4633980406386\\
0.0220000000000056	92.2056422351234\\
0.0229999999999961	92.8424346879535\\
0.0240000000000009	93.4148222060958\\
0.0250000000000057	93.9420809143686\\
0.0259999999999962	94.4251179245283\\
0.027000000000001	94.848965892598\\
0.0280000000000058	95.2267779390421\\
0.0289999999999964	95.5789640783745\\
0.0300000000000011	95.8980406386067\\
0.0310000000000059	96.1860486211901\\
0.0319999999999965	96.4484306966618\\
0.0330000000000013	96.6962989840348\\
0.034000000000006	96.9035740203193\\
0.0349999999999966	97.079553701016\\
0.0360000000000014	97.2621099419449\\
0.0390000000000015	97.7251904934688\\
0.0400000000000063	97.8621643686502\\
0.0409999999999968	97.9809960087083\\
0.0420000000000016	98.092343976778\\
0.0430000000000064	98.1932601596517\\
0.0439999999999969	98.2984851233672\\
0.0450000000000017	98.3905569666183\\
0.046999999999997	98.5581458635704\\
0.0480000000000018	98.6357039187228\\
0.0499999999999972	98.7962626995646\\
0.0510000000000019	98.8654299709724\\
0.0520000000000067	98.9203102322206\\
0.0529999999999973	98.9722423802612\\
0.0559999999999974	99.1146589259797\\
0.0570000000000022	99.1516237300435\\
0.0600000000000023	99.2804335994195\\
0.0619999999999976	99.3491473149492\\
0.0630000000000024	99.385658563135\\
0.063999999999993	99.4285195936139\\
0.0649999999999977	99.4616291727141\\
0.0660000000000025	99.4899764150943\\
0.0669999999999931	99.5156023222061\\
0.0679999999999978	99.5400943396226\\
0.0690000000000026	99.563679245283\\
0.0699999999999932	99.59043904209\\
0.070999999999998	99.6187862844702\\
0.0720000000000027	99.6416908563135\\
0.0729999999999933	99.660740203193\\
0.0750000000000028	99.7074564586357\\
0.0759999999999934	99.7326288098694\\
0.0769999999999982	99.7512246008708\\
0.078000000000003	99.7707275036284\\
0.0789999999999935	99.7852412917271\\
0.0799999999999983	99.7945391872279\\
0.0810000000000031	99.8072387518142\\
0.0819999999999936	99.8140420899855\\
0.0829999999999984	99.8201650943396\\
0.0840000000000032	99.8258345428157\\
0.0859999999999985	99.8376269956459\\
0.0870000000000033	99.8426161103048\\
0.0879999999999939	99.8498730043541\\
0.0889999999999986	99.8530478955007\\
0.0900000000000034	99.8578102322206\\
0.090999999999994	99.8627993468795\\
0.0919999999999987	99.8668813497823\\
0.0930000000000035	99.870736574746\\
0.0939999999999941	99.8743650217707\\
0.0949999999999989	99.8793541364296\\
0.097999999999999	99.8888788098694\\
0.0990000000000038	99.8913733671989\\
0.0999999999999943	99.8934143686502\\
0.100999999999999	99.8968160377358\\
};
\addlegendentry{\footnotesize \textbf{Ours}}

\end{axis}
\end{tikzpicture}%

%% file: scape.tikz
%
%
\definecolor{mycolor1}{rgb}{0.00000,0.44700,0.74100}%
\definecolor{mycolor2}{rgb}{0.85000,0.32500,0.09800}%
\definecolor{mycolor3}{rgb}{0.92900,0.69400,0.12500}%
\pgfplotsset{scaled y ticks=false}
\pgfplotsset{scaled x ticks=false}
\begin{tikzpicture}

\begin{axis}[%
width=0.6\columnwidth,
height=0.42\columnwidth,
scale only axis,
xmin=0,
xmax=0.1,
xlabel style={font=\color{white!15!black}},
xlabel={\footnotesize Geodesic error},
xlabel style={at={(0.5,0.05)}},
ymin=0,
ymax=1,
yticklabels={0,20,40,60,80,100},
ytick={0,0.2,0.4,0.6,0.8,1.0},
xtick={0,0.02,0.04,0.06,0.08,0.1},
xticklabels={0,0.02,0.04,0.06,0.08,0.1},
ylabel style={font=\color{white!15!black}},
ylabel={\footnotesize \% Correspondences},
ylabel style={at={(0.12,0.48)}},
every x tick label/.append style={font=\color{black}, font=\tiny},
every y tick label/.append style={font=\color{black}, font=\tiny},
axis background/.style={fill=white},
axis x line*=bottom,
axis y line*=left,
xmajorgrids,
ymajorgrids,
legend style={at={(1.03,0.16)}, anchor=south west, legend cell align=left, align=left, draw=white!15!black},
title={\footnotesize \textbf{SCAPE}},
title style={at={(0.5,0.96)}}
]
\addplot [color=mycolor1, line width=2.0pt]
  table[row sep=crcr]{%
0	0.2134\\
0.001	0.213448\\
0.002	0.215056\\
0.003	0.22456\\
0.004	0.250824\\
0.005	0.292736\\
0.00600000000000001	0.344128\\
0.00700000000000001	0.38988\\
0.00800000000000001	0.42572\\
0.00900000000000001	0.452856\\
0.01	0.475272\\
0.011	0.495712\\
0.012	0.5152\\
0.013	0.532504\\
0.014	0.548384\\
0.015	0.563464\\
0.016	0.577064\\
0.017	0.5896\\
0.019	0.6126\\
0.022	0.645024\\
0.025	0.674768\\
0.026	0.683048\\
0.027	0.690864\\
0.029	0.705688\\
0.03	0.712712\\
0.032	0.72592\\
0.034	0.738616\\
0.035	0.744656\\
0.036	0.750248\\
0.039	0.767664\\
0.042	0.783352\\
0.045	0.798232\\
0.046	0.803072\\
0.047	0.807632\\
0.048	0.811992\\
0.049	0.81664\\
0.05	0.821048\\
0.051	0.825256\\
0.052	0.829304\\
0.053	0.833\\
0.054	0.837088\\
0.055	0.84104\\
0.056	0.844704\\
0.0570000000000001	0.848232\\
0.0580000000000001	0.851984\\
0.0590000000000001	0.855056\\
0.0600000000000001	0.85852\\
0.0610000000000001	0.861848\\
0.0640000000000001	0.87132\\
0.0650000000000001	0.874224\\
0.0680000000000001	0.88192\\
0.0690000000000001	0.88428\\
0.073	0.893184\\
0.074	0.895128\\
0.075	0.896912\\
0.076	0.898848\\
0.077	0.900856\\
0.078	0.902648\\
0.079	0.904576\\
0.083	0.911488\\
0.084	0.912952\\
0.085	0.91472\\
0.086	0.916296\\
0.088	0.919032\\
0.089	0.920544\\
0.09	0.92216\\
0.092	0.92492\\
0.093	0.926232\\
0.095	0.928728\\
0.096	0.930128\\
0.097	0.931304\\
0.099	0.933552\\
0.101	0.935576\\
};
\addlegendentry{\footnotesize \textbf{Ours}}

\addplot [color=black,solid,line width=1.0pt]
  table[row sep=crcr]{%
0	0.131128888888889\\
0.001	0.131128888888889\\
0.002	0.133084444444444\\
0.003	0.145804444444444\\
0.004	0.181511111111111\\
0.005	0.241235555555556\\
0.006	0.313635555555556\\
0.007	0.379395555555556\\
0.008	0.437413333333333\\
0.009	0.486746666666667\\
0.01	0.531271111111111\\
0.011	0.572488888888889\\
0.012	0.608106666666667\\
0.013	0.640266666666667\\
0.014	0.668506666666667\\
0.015	0.693208888888889\\
0.016	0.714568888888889\\
0.017	0.732986666666667\\
0.018	0.749475555555556\\
0.019	0.764995555555556\\
0.02	0.778737777777778\\
0.021	0.791431111111111\\
0.022	0.803555555555556\\
0.023	0.813413333333333\\
0.024	0.822968888888889\\
0.025	0.831413333333333\\
0.026	0.838977777777778\\
0.027	0.846213333333333\\
0.028	0.853217777777778\\
0.029	0.859653333333333\\
0.03	0.865866666666667\\
0.031	0.871662222222222\\
0.032	0.877075555555556\\
0.033	0.882417777777778\\
0.034	0.887333333333333\\
0.035	0.891733333333333\\
0.036	0.89616\\
0.037	0.900533333333333\\
0.038	0.904506666666667\\
0.039	0.908577777777778\\
0.04	0.912817777777778\\
0.041	0.916604444444444\\
0.042	0.920684444444444\\
0.043	0.92424\\
0.044	0.927413333333333\\
0.045	0.930311111111111\\
0.046	0.933048888888889\\
0.047	0.935955555555556\\
0.048	0.938657777777778\\
0.049	0.941208888888889\\
0.05	0.943893333333333\\
0.051	0.94616\\
0.052	0.948266666666667\\
0.053	0.950524444444445\\
0.054	0.952684444444445\\
0.055	0.954711111111111\\
0.056	0.956613333333333\\
0.057	0.958497777777778\\
0.058	0.959991111111111\\
0.059	0.961475555555556\\
0.06	0.962648888888889\\
0.061	0.963884444444444\\
0.062	0.965093333333333\\
0.063	0.96608\\
0.064	0.967004444444444\\
0.065	0.967795555555556\\
0.066	0.96856\\
0.067	0.969244444444444\\
0.068	0.969644444444444\\
0.069	0.970088888888889\\
0.07	0.970346666666667\\
0.071	0.970568888888889\\
0.072	0.970817777777778\\
0.073	0.971013333333333\\
0.074	0.971235555555556\\
0.075	0.971368888888889\\
0.076	0.971528888888889\\
0.077	0.971617777777778\\
0.078	0.97176\\
0.079	0.971848888888889\\
0.08	0.971937777777778\\
0.081	0.972088888888889\\
0.082	0.972195555555556\\
0.083	0.972302222222222\\
0.084	0.972417777777778\\
0.085	0.972497777777778\\
0.086	0.972568888888889\\
0.087	0.972657777777778\\
0.088	0.9728\\
0.089	0.972915555555555\\
0.09	0.97304\\
0.091	0.973226666666666\\
0.092	0.973297777777778\\
0.093	0.973422222222222\\
0.094	0.973537777777778\\
0.095	0.973644444444444\\
0.096	0.973688888888889\\
0.097	0.973777777777778\\
0.098	0.973866666666667\\
0.099	0.973955555555555\\
0.1	0.97408\\
0.101	0.974195555555555\\
0.102	0.974328888888889\\
0.103	0.974506666666667\\
0.104	0.974675555555555\\
0.105	0.974826666666667\\
0.106	0.975048888888889\\
0.107	0.975173333333333\\
0.108	0.975262222222222\\
0.109	0.975413333333333\\
0.11	0.975582222222222\\
0.111	0.975733333333333\\
0.112	0.97592\\
0.113	0.976035555555555\\
0.114	0.97624\\
0.115	0.976346666666667\\
0.116	0.976435555555556\\
0.117	0.976542222222222\\
0.118	0.976657777777778\\
0.119	0.976728888888889\\
0.12	0.976844444444444\\
0.121	0.977004444444444\\
0.122	0.977137777777778\\
0.123	0.977262222222222\\
0.124	0.977333333333333\\
0.125	0.977422222222222\\
0.126	0.977502222222222\\
0.127	0.977608888888889\\
0.128	0.977688888888889\\
0.129	0.977751111111111\\
0.13	0.977813333333333\\
0.131	0.977893333333333\\
0.132	0.977955555555555\\
0.133	0.978053333333333\\
0.134	0.978142222222222\\
0.135	0.978213333333333\\
0.136	0.978257777777778\\
0.137	0.978266666666667\\
0.138	0.978275555555555\\
0.139	0.978284444444444\\
0.14	0.978284444444444\\
0.141	0.978284444444444\\
0.142	0.978284444444444\\
0.143	0.978284444444444\\
0.144	0.978284444444444\\
0.145	0.978284444444444\\
0.146	0.978284444444444\\
0.147	0.978284444444444\\
0.148	0.978284444444444\\
0.149	0.978284444444444\\
0.15	0.978284444444444\\
0.151	0.978284444444444\\
0.152	0.978284444444444\\
0.153	0.978284444444444\\
0.154	0.978284444444444\\
0.155	0.978293333333333\\
0.156	0.978302222222222\\
0.157	0.978302222222222\\
0.158	0.978302222222222\\
0.159	0.978302222222222\\
0.16	0.978311111111111\\
0.161	0.97832\\
0.162	0.978328888888889\\
0.163	0.978337777777778\\
0.164	0.978337777777778\\
0.165	0.978337777777778\\
0.166	0.978346666666667\\
0.167	0.978346666666667\\
0.168	0.978364444444444\\
0.169	0.978373333333333\\
0.17	0.978382222222222\\
0.171	0.978382222222222\\
0.172	0.978382222222222\\
0.173	0.978391111111111\\
0.174	0.9784\\
0.175	0.978408888888889\\
0.176	0.978435555555556\\
0.177	0.978462222222222\\
0.178	0.978488888888889\\
0.179	0.978542222222222\\
0.18	0.978595555555556\\
0.181	0.978613333333333\\
0.182	0.97864\\
0.183	0.978675555555556\\
0.184	0.978693333333333\\
0.185	0.97872\\
0.186	0.978737777777778\\
0.187	0.978764444444444\\
0.188	0.978782222222222\\
0.189	0.978782222222222\\
0.19	0.978808888888889\\
0.191	0.978826666666667\\
0.192	0.978862222222222\\
0.193	0.978871111111111\\
0.194	0.978897777777778\\
0.195	0.978897777777778\\
0.196	0.978906666666667\\
0.197	0.978924444444444\\
0.198	0.978942222222222\\
0.199	0.978942222222222\\
0.2	0.978942222222222\\
0.201	0.978942222222222\\
0.202	0.978951111111111\\
0.203	0.978951111111111\\
0.204	0.978951111111111\\
0.205	0.978951111111111\\
0.206	0.978951111111111\\
0.207	0.978951111111111\\
0.208	0.978951111111111\\
0.209	0.978951111111111\\
0.21	0.978951111111111\\
0.211	0.978951111111111\\
0.212	0.978951111111111\\
0.213	0.978951111111111\\
0.214	0.978951111111111\\
0.215	0.97896\\
0.216	0.978968888888889\\
0.217	0.978977777777778\\
0.218	0.978977777777778\\
0.219	0.978986666666667\\
0.22	0.978986666666667\\
0.221	0.978995555555556\\
0.222	0.978995555555556\\
0.223	0.978995555555556\\
0.224	0.978995555555556\\
0.225	0.978995555555556\\
0.226	0.978995555555556\\
0.227	0.978995555555556\\
0.228	0.978995555555556\\
0.229	0.978995555555556\\
0.23	0.978995555555556\\
0.231	0.978995555555556\\
0.232	0.978995555555556\\
0.233	0.978995555555556\\
0.234	0.978995555555556\\
0.235	0.978995555555556\\
0.236	0.978995555555556\\
0.237	0.978995555555556\\
0.238	0.978995555555556\\
0.239	0.978995555555556\\
0.24	0.978995555555556\\
0.241	0.978995555555556\\
0.242	0.978995555555556\\
0.243	0.978995555555556\\
0.244	0.978995555555556\\
0.245	0.978995555555556\\
0.246	0.978995555555556\\
0.247	0.978995555555556\\
0.248	0.978995555555556\\
0.249	0.978995555555556\\
0.25	0.978995555555556\\
0.251	0.979004444444444\\
0.252	0.979004444444444\\
0.253	0.979013333333333\\
0.254	0.979022222222222\\
0.255	0.979022222222222\\
0.256	0.97904\\
0.257	0.979057777777778\\
0.258	0.979057777777778\\
0.259	0.979066666666667\\
0.26	0.979066666666667\\
0.261	0.979075555555556\\
0.262	0.979093333333333\\
0.263	0.979111111111111\\
0.264	0.97912\\
0.265	0.979128888888889\\
0.266	0.979128888888889\\
0.267	0.979137777777778\\
0.268	0.979137777777778\\
0.269	0.979137777777778\\
0.27	0.979146666666667\\
0.271	0.979146666666667\\
0.272	0.979146666666667\\
0.273	0.979146666666667\\
0.274	0.979146666666667\\
0.275	0.979146666666667\\
0.276	0.979146666666667\\
0.277	0.979146666666667\\
0.278	0.979146666666667\\
0.279	0.979146666666667\\
0.28	0.979146666666667\\
0.281	0.979146666666667\\
0.282	0.979146666666667\\
0.283	0.979146666666667\\
0.284	0.979146666666667\\
0.285	0.979146666666667\\
0.286	0.979146666666667\\
0.287	0.979146666666667\\
0.288	0.979146666666667\\
0.289	0.979146666666667\\
0.29	0.979146666666667\\
0.291	0.979146666666667\\
0.292	0.979146666666667\\
0.293	0.979146666666667\\
0.294	0.979146666666667\\
0.295	0.979146666666667\\
0.296	0.979146666666667\\
0.297	0.979146666666667\\
0.298	0.979146666666667\\
0.299	0.979146666666667\\
0.3	0.979146666666667\\
0.301	0.979146666666667\\
};
\addlegendentry{\footnotesize FMNet~\cite{litany2017deep}}

\addplot [color=mycolor2,solid,line width=1.25pt]
  table[row sep=crcr]{%
0	0.0289777777777778\\
0.003	0.02952\\
0.006	0.0437777777777778\\
0.009	0.0845422222222222\\
0.012	0.129075555555556\\
0.015	0.166071111111111\\
0.018	0.201137777777778\\
0.021	0.237671111111111\\
0.024	0.269608888888889\\
0.027	0.29776\\
0.03	0.325404444444445\\
0.033	0.350053333333333\\
0.036	0.372995555555556\\
0.039	0.393733333333333\\
0.042	0.414142222222222\\
0.045	0.4344\\
0.048	0.453013333333333\\
0.051	0.471022222222222\\
0.054	0.488213333333333\\
0.057	0.504773333333333\\
0.06	0.520977777777778\\
0.063	0.536924444444444\\
0.066	0.551635555555556\\
0.069	0.565564444444444\\
0.072	0.579022222222222\\
0.075	0.592133333333333\\
0.078	0.604604444444445\\
0.081	0.616844444444444\\
0.084	0.628168888888889\\
0.087	0.638622222222222\\
0.09	0.649324444444444\\
0.093	0.659173333333333\\
0.096	0.668453333333333\\
0.099	0.677297777777778\\
0.102	0.686088888888889\\
0.105	0.693493333333333\\
0.108	0.700968888888889\\
0.111	0.707742222222222\\
0.114	0.713893333333333\\
0.117	0.719511111111111\\
0.12	0.725564444444445\\
0.123	0.731235555555556\\
0.126	0.736835555555556\\
0.129	0.742328888888889\\
0.132	0.747742222222222\\
0.135	0.752728888888889\\
0.138	0.757315555555556\\
0.141	0.761875555555556\\
0.144	0.765555555555556\\
0.147	0.769217777777778\\
0.15	0.772924444444444\\
0.153	0.7764\\
0.156	0.779831111111111\\
0.159	0.782951111111111\\
0.162	0.786275555555556\\
0.165	0.78976\\
0.168	0.792693333333333\\
0.171	0.795973333333333\\
0.174	0.799111111111111\\
0.177	0.802453333333333\\
0.18	0.805448888888889\\
0.183	0.808826666666667\\
0.186	0.811831111111111\\
0.189	0.814693333333333\\
0.192	0.817626666666667\\
0.195	0.820328888888889\\
0.198	0.822951111111111\\
0.201	0.825857777777778\\
0.204	0.828364444444444\\
0.207	0.830862222222222\\
0.21	0.833466666666667\\
0.213	0.836017777777778\\
0.216	0.838675555555556\\
0.219	0.841022222222222\\
0.222	0.843813333333333\\
0.225	0.846204444444445\\
0.228	0.848951111111111\\
0.231	0.851653333333333\\
0.234	0.854071111111111\\
0.237	0.856168888888889\\
0.24	0.858222222222222\\
0.243	0.860337777777778\\
0.246	0.862337777777778\\
0.249	0.864328888888889\\
0.252	0.866346666666667\\
0.255	0.868177777777778\\
0.258	0.87\\
0.261	0.87192\\
0.264	0.873493333333333\\
0.267	0.875333333333333\\
0.27	0.877137777777778\\
0.273	0.878773333333333\\
0.276	0.880453333333333\\
0.279	0.88192\\
0.282	0.883644444444445\\
0.285	0.885137777777778\\
0.288	0.886746666666667\\
0.291	0.888382222222222\\
0.294	0.890088888888889\\
0.297	0.891733333333333\\
0.3	0.893395555555556\\
};
\addlegendentry{\footnotesize OSD~\cite{osd}};

\addplot [color=black!40!green,solid,line width=1.25pt]
  table[row sep=crcr]{%
0	0.0329955555555556\\
0.003	0.0337244444444444\\
0.006	0.0460622222222222\\
0.009	0.0856444444444444\\
0.012	0.135013333333333\\
0.015	0.179182222222222\\
0.018	0.223208888888889\\
0.021	0.268115555555556\\
0.024	0.311946666666667\\
0.027	0.352275555555556\\
0.03	0.388426666666667\\
0.033	0.422231111111111\\
0.036	0.453742222222222\\
0.039	0.482551111111111\\
0.042	0.509431111111111\\
0.045	0.534986666666667\\
0.048	0.558124444444444\\
0.051	0.579848888888889\\
0.054	0.59968\\
0.057	0.619164444444445\\
0.06	0.63776\\
0.063	0.655928888888889\\
0.066	0.672328888888889\\
0.069	0.688675555555555\\
0.072	0.70352\\
0.075	0.718177777777778\\
0.078	0.732897777777778\\
0.081	0.747386666666667\\
0.084	0.76048\\
0.087	0.773164444444445\\
0.09	0.784684444444444\\
0.093	0.79512\\
0.096	0.805324444444444\\
0.099	0.815128888888889\\
0.102	0.824506666666667\\
0.105	0.833173333333333\\
0.108	0.841173333333333\\
0.111	0.848168888888889\\
0.114	0.854942222222222\\
0.117	0.86144\\
0.12	0.867804444444444\\
0.123	0.874231111111111\\
0.126	0.879831111111111\\
0.129	0.885111111111111\\
0.132	0.890142222222222\\
0.135	0.89496\\
0.138	0.899653333333333\\
0.141	0.9036\\
0.144	0.907093333333333\\
0.147	0.910035555555555\\
0.15	0.913191111111111\\
0.153	0.916346666666667\\
0.156	0.919262222222222\\
0.159	0.922195555555555\\
0.162	0.924782222222223\\
0.165	0.92768\\
0.168	0.930231111111111\\
0.171	0.93272\\
0.174	0.93496\\
0.177	0.93736\\
0.18	0.939715555555556\\
0.183	0.942044444444444\\
0.186	0.944204444444444\\
0.189	0.946453333333333\\
0.192	0.948533333333333\\
0.195	0.950782222222222\\
0.198	0.952764444444445\\
0.201	0.954568888888889\\
0.204	0.956533333333333\\
0.207	0.958142222222222\\
0.21	0.960017777777778\\
0.213	0.961902222222222\\
0.216	0.963493333333333\\
0.219	0.965271111111111\\
0.222	0.966737777777778\\
0.225	0.968044444444444\\
0.228	0.969582222222222\\
0.231	0.971164444444444\\
0.234	0.972533333333333\\
0.237	0.973822222222222\\
0.24	0.974986666666667\\
0.243	0.976275555555556\\
0.246	0.977555555555556\\
0.249	0.978524444444445\\
0.252	0.979546666666667\\
0.255	0.980311111111111\\
0.258	0.981084444444444\\
0.261	0.981902222222222\\
0.264	0.982906666666667\\
0.267	0.983644444444444\\
0.27	0.9844\\
0.273	0.985084444444444\\
0.276	0.985768888888889\\
0.279	0.986444444444444\\
0.282	0.987004444444444\\
0.285	0.9876\\
0.288	0.988195555555556\\
0.291	0.988782222222222\\
0.294	0.989253333333333\\
0.297	0.989697777777778\\
0.3	0.990115555555556\\
};
\addlegendentry{\footnotesize GCNN~\cite{masci2015geodesic}}; 

\addplot [color=mycolor3,solid,line width=1.25pt]
  table[row sep=crcr]{%
0	0.0359733333333333\\
0.003	0.0362933333333333\\
0.006	0.0485066666666667\\
0.009	0.0995733333333333\\
0.012	0.1616\\
0.015	0.21528\\
0.018	0.266071111111111\\
0.021	0.319297777777778\\
0.024	0.367768888888889\\
0.027	0.41144\\
0.03	0.451173333333333\\
0.033	0.486106666666667\\
0.036	0.517475555555556\\
0.039	0.545582222222222\\
0.042	0.572257777777778\\
0.045	0.596257777777778\\
0.048	0.618497777777778\\
0.051	0.640195555555556\\
0.054	0.659137777777778\\
0.057	0.677742222222222\\
0.06	0.695146666666667\\
0.063	0.711822222222222\\
0.066	0.727848888888889\\
0.069	0.743431111111111\\
0.072	0.757528888888889\\
0.075	0.771244444444445\\
0.078	0.784248888888889\\
0.081	0.796542222222222\\
0.084	0.808266666666667\\
0.087	0.818728888888889\\
0.09	0.829475555555556\\
0.093	0.838471111111111\\
0.096	0.846506666666667\\
0.099	0.854746666666667\\
0.102	0.862062222222222\\
0.105	0.868515555555555\\
0.108	0.87472\\
0.111	0.880488888888889\\
0.114	0.885795555555556\\
0.117	0.890435555555556\\
0.12	0.895431111111111\\
0.123	0.900124444444444\\
0.126	0.904382222222222\\
0.129	0.908524444444444\\
0.132	0.912355555555556\\
0.135	0.915866666666667\\
0.138	0.918808888888889\\
0.141	0.921822222222222\\
0.144	0.924355555555555\\
0.147	0.926675555555555\\
0.15	0.928888888888889\\
0.153	0.931084444444444\\
0.156	0.933164444444444\\
0.159	0.935244444444445\\
0.162	0.937217777777778\\
0.165	0.939244444444445\\
0.168	0.940684444444444\\
0.171	0.942577777777778\\
0.174	0.944186666666667\\
0.177	0.945911111111111\\
0.18	0.947377777777778\\
0.183	0.94928\\
0.186	0.950764444444445\\
0.189	0.952346666666667\\
0.192	0.954177777777778\\
0.195	0.955768888888889\\
0.198	0.957146666666667\\
0.201	0.958675555555555\\
0.204	0.960168888888889\\
0.207	0.961493333333333\\
0.21	0.962666666666667\\
0.213	0.964133333333333\\
0.216	0.965528888888889\\
0.219	0.966844444444445\\
0.222	0.968177777777778\\
0.225	0.969653333333333\\
0.228	0.970684444444444\\
0.231	0.971955555555556\\
0.234	0.97304\\
0.237	0.974328888888889\\
0.24	0.975466666666667\\
0.243	0.976506666666667\\
0.246	0.977564444444444\\
0.249	0.978773333333333\\
0.252	0.979831111111111\\
0.255	0.980791111111111\\
0.258	0.981902222222222\\
0.261	0.983066666666667\\
0.264	0.984053333333333\\
0.267	0.985217777777778\\
0.27	0.986097777777778\\
0.273	0.986933333333333\\
0.276	0.987875555555556\\
0.279	0.988728888888889\\
0.282	0.989653333333334\\
0.285	0.990506666666667\\
0.288	0.991288888888889\\
0.291	0.992346666666667\\
0.294	0.993075555555556\\
0.297	0.993964444444444\\
0.3	0.994595555555555\\
};
\addlegendentry{\footnotesize LSCNN~\cite{WFT2015}}; 

\addplot [color=black!20!blue,solid,line width=1.25pt]
  table[row sep=crcr]{%
0	0.0297155555555556\\
0.003	0.0301955555555556\\
0.006	0.0427377777777778\\
0.009	0.0857511111111111\\
0.012	0.137884444444444\\
0.015	0.184222222222222\\
0.018	0.230968888888889\\
0.021	0.278915555555556\\
0.024	0.323084444444444\\
0.027	0.363662222222222\\
0.03	0.401342222222222\\
0.033	0.436044444444444\\
0.036	0.467893333333333\\
0.039	0.496675555555556\\
0.042	0.525164444444445\\
0.045	0.5508\\
0.048	0.574488888888889\\
0.051	0.598142222222222\\
0.054	0.619075555555555\\
0.057	0.639688888888889\\
0.06	0.659413333333333\\
0.063	0.678186666666667\\
0.066	0.695813333333333\\
0.069	0.712853333333333\\
0.072	0.728462222222222\\
0.075	0.743271111111111\\
0.078	0.757866666666667\\
0.081	0.771093333333333\\
0.084	0.783928888888889\\
0.087	0.795546666666667\\
0.09	0.806817777777778\\
0.093	0.81752\\
0.096	0.827306666666667\\
0.099	0.836417777777778\\
0.102	0.845111111111111\\
0.105	0.85248\\
0.108	0.858924444444444\\
0.111	0.865288888888889\\
0.114	0.871066666666667\\
0.117	0.876195555555556\\
0.12	0.881217777777778\\
0.123	0.885937777777778\\
0.126	0.890444444444444\\
0.129	0.894702222222222\\
0.132	0.898711111111111\\
0.135	0.902328888888889\\
0.138	0.905715555555556\\
0.141	0.90904\\
0.144	0.912177777777778\\
0.147	0.914924444444444\\
0.15	0.917484444444444\\
0.153	0.920062222222222\\
0.156	0.922471111111111\\
0.159	0.924933333333333\\
0.162	0.927208888888889\\
0.165	0.929626666666667\\
0.168	0.932195555555555\\
0.171	0.934186666666667\\
0.174	0.936213333333333\\
0.177	0.938311111111111\\
0.18	0.940204444444445\\
0.183	0.942186666666667\\
0.186	0.944106666666667\\
0.189	0.945848888888889\\
0.192	0.9476\\
0.195	0.949386666666667\\
0.198	0.950942222222222\\
0.201	0.952711111111111\\
0.204	0.95448\\
0.207	0.956133333333333\\
0.21	0.957946666666667\\
0.213	0.959813333333333\\
0.216	0.961422222222222\\
0.219	0.963057777777778\\
0.222	0.964622222222222\\
0.225	0.966284444444444\\
0.228	0.967671111111111\\
0.231	0.969182222222222\\
0.234	0.970622222222222\\
0.237	0.972026666666667\\
0.24	0.97336\\
0.243	0.974631111111111\\
0.246	0.975822222222222\\
0.249	0.977004444444444\\
0.252	0.978328888888889\\
0.255	0.979715555555555\\
0.258	0.980897777777778\\
0.261	0.981884444444444\\
0.264	0.982853333333333\\
0.267	0.983866666666667\\
0.27	0.984666666666667\\
0.273	0.985591111111111\\
0.276	0.986604444444445\\
0.279	0.98744\\
0.282	0.988302222222222\\
0.285	0.989253333333333\\
0.288	0.990133333333333\\
0.291	0.99096\\
0.294	0.991635555555555\\
0.297	0.992195555555555\\
0.3	0.992853333333333\\
};
\addlegendentry{\footnotesize ADD3~\cite{add16}} 

\addplot [color=black!20!red,solid,line width=1.25pt]
  table[row sep=crcr]{%
0	0.0327022222222222\\
0.003	0.0333066666666667\\
0.006	0.04792\\
0.009	0.0950488888888889\\
0.012	0.151404444444444\\
0.015	0.200026666666667\\
0.018	0.249733333333333\\
0.021	0.299502222222222\\
0.024	0.345457777777778\\
0.027	0.388497777777778\\
0.03	0.427173333333333\\
0.033	0.462808888888889\\
0.036	0.496231111111111\\
0.039	0.526577777777778\\
0.042	0.554915555555556\\
0.045	0.579831111111111\\
0.048	0.604862222222222\\
0.051	0.628702222222222\\
0.054	0.650151111111111\\
0.057	0.670506666666667\\
0.06	0.689191111111111\\
0.063	0.706488888888889\\
0.066	0.723315555555556\\
0.069	0.738435555555556\\
0.072	0.75288\\
0.075	0.765573333333333\\
0.078	0.778222222222222\\
0.081	0.790008888888889\\
0.084	0.801653333333333\\
0.087	0.812026666666667\\
0.09	0.821511111111111\\
0.093	0.830604444444444\\
0.096	0.83912\\
0.099	0.846764444444445\\
0.102	0.85352\\
0.105	0.859653333333333\\
0.108	0.865528888888889\\
0.111	0.870444444444444\\
0.114	0.875253333333333\\
0.117	0.879946666666667\\
0.12	0.884026666666667\\
0.123	0.887991111111111\\
0.126	0.892071111111111\\
0.129	0.895608888888889\\
0.132	0.899102222222222\\
0.135	0.902488888888889\\
0.138	0.9056\\
0.141	0.908595555555556\\
0.144	0.911271111111111\\
0.147	0.913964444444444\\
0.15	0.91664\\
0.153	0.919173333333333\\
0.156	0.921253333333333\\
0.159	0.923262222222222\\
0.162	0.925226666666667\\
0.165	0.927537777777778\\
0.168	0.929466666666667\\
0.171	0.931395555555556\\
0.174	0.933333333333333\\
0.177	0.935324444444444\\
0.18	0.937208888888889\\
0.183	0.939111111111111\\
0.186	0.941048888888889\\
0.189	0.942906666666667\\
0.192	0.944791111111111\\
0.195	0.946648888888889\\
0.198	0.948417777777778\\
0.201	0.95016\\
0.204	0.9516\\
0.207	0.953102222222222\\
0.21	0.954542222222222\\
0.213	0.956151111111111\\
0.216	0.957448888888889\\
0.219	0.958933333333333\\
0.222	0.960613333333333\\
0.225	0.962151111111111\\
0.228	0.963822222222222\\
0.231	0.965244444444445\\
0.234	0.966684444444444\\
0.237	0.968124444444444\\
0.24	0.969564444444444\\
0.243	0.970986666666667\\
0.246	0.972257777777778\\
0.249	0.973724444444444\\
0.252	0.975155555555556\\
0.255	0.976462222222222\\
0.258	0.977937777777778\\
0.261	0.979431111111111\\
0.264	0.98104\\
0.267	0.982631111111111\\
0.27	0.984053333333333\\
0.273	0.985351111111111\\
0.276	0.986817777777778\\
0.279	0.987795555555556\\
0.282	0.989031111111111\\
0.285	0.989964444444444\\
0.288	0.991057777777778\\
0.291	0.991875555555556\\
0.294	0.992817777777778\\
0.297	0.993564444444444\\
0.3	0.9944\\
};
\addlegendentry{\footnotesize mADD3 \cite{add16}}; 

\end{axis}
\end{tikzpicture}%

%% file: appendix.tex
\begin{figure*}[t]
\begin{center}
    \centering
    \begin{overpic}
    [trim=0cm 0cm 0cm 0cm,clip,width=0.212\linewidth]{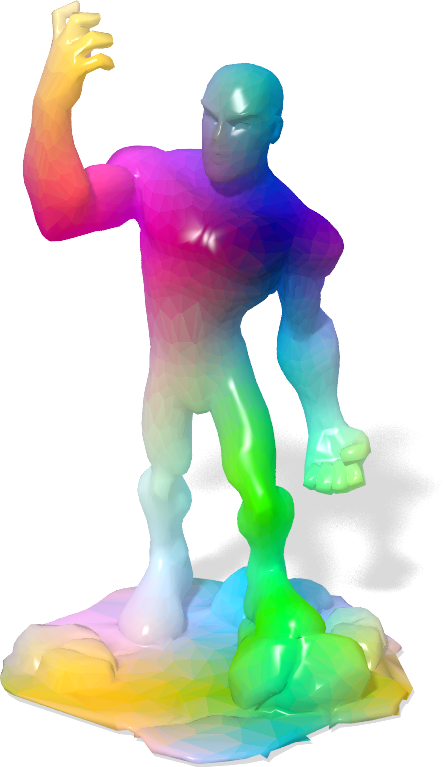}
    \put(28,95){\footnotesize Ref.}
    \end{overpic}
    \begin{overpic}
    [trim=0cm 0cm 0cm 0cm,clip,width=0.25\linewidth]{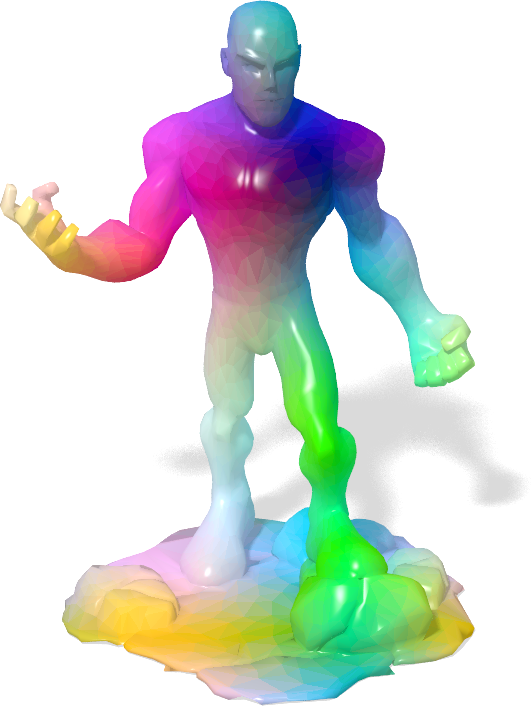}
    \put(32,105){\footnotesize \textbf{Ours}}
    \end{overpic}
    \begin{overpic}
    [trim=0cm 0cm 0cm 0cm,clip,width=0.25\linewidth]{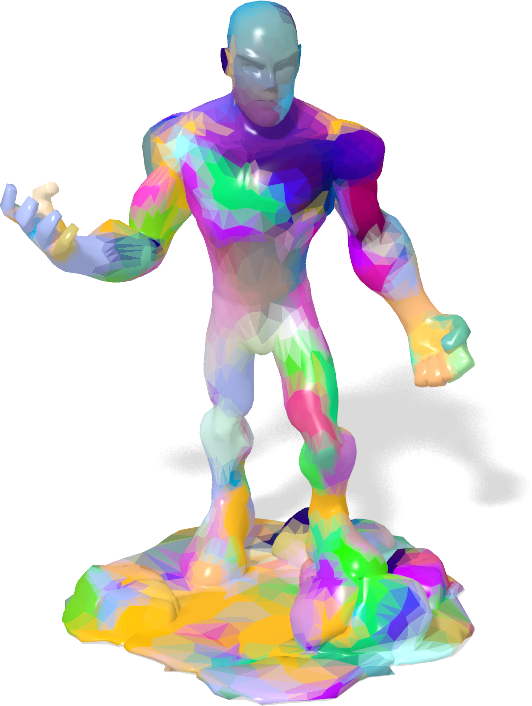}
    \put(28,105){\footnotesize FMNet~\cite{litany2017deep}}
    \end{overpic}
    \begin{overpic}
    [trim=0cm 0cm 0cm 0cm,clip,width=0.25\linewidth]{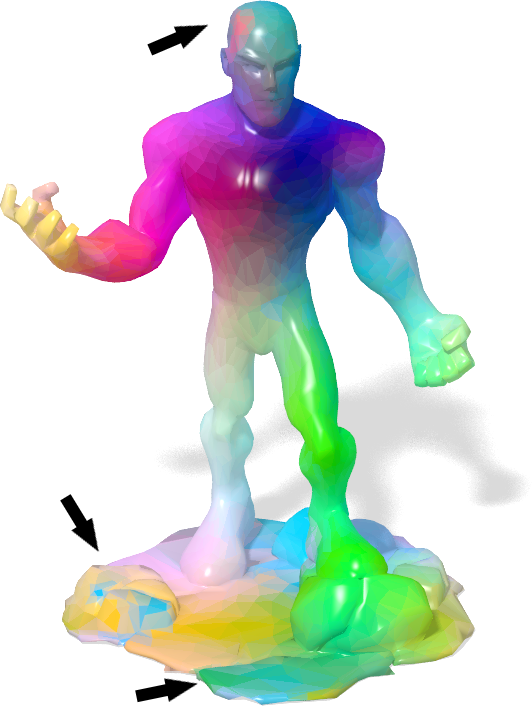}
    \put(30,105){\footnotesize FM~\cite{ovsjanikov12}}
    \end{overpic}
    \vspace{0.1cm}
\caption{\label{fig:hero}Single-pair correspondences obtained with our unsupervised scheme, a state-of-the-art supervised approach FMNet~\cite{litany2017deep}, and an axiomatic matching pipeline. Our approach was trained on the single pair shown in this example (where point-to-point correspondence was {\em not} manually annotated); FMNet was trained on humans from the FAUST dataset~\cite{faust}.}
\end{center}%
\end{figure*}

\begin{figure}[t]
    \centering
    \includegraphics[width=0.71\linewidth]{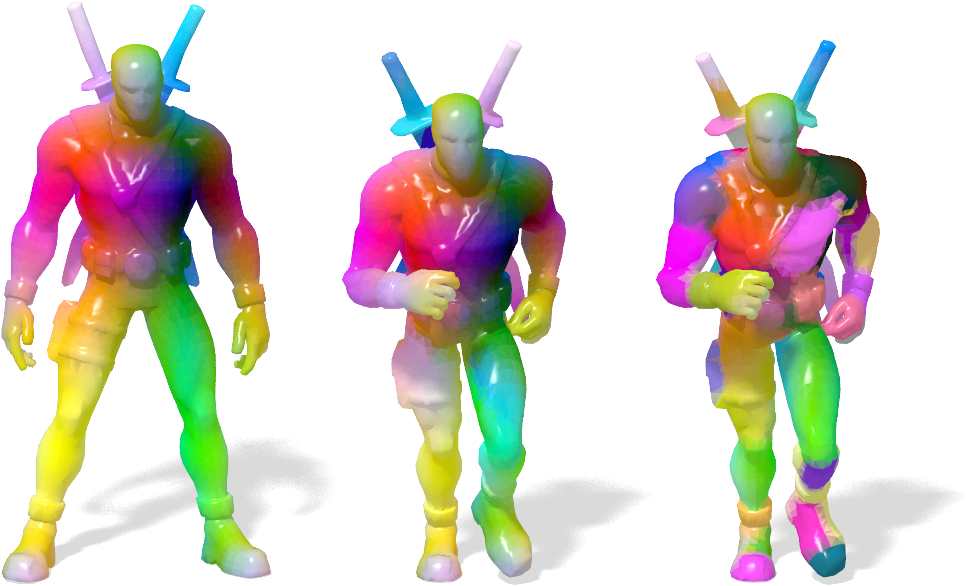}
    \includegraphics[width=0.24\linewidth]{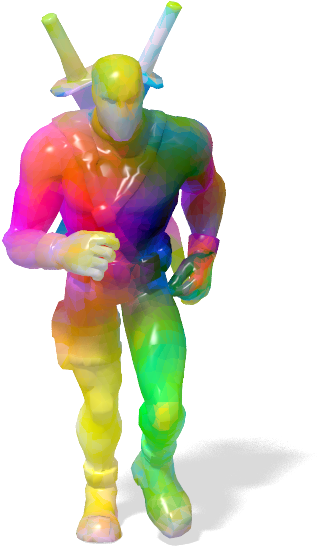}
    \includegraphics[width=0.7\linewidth]{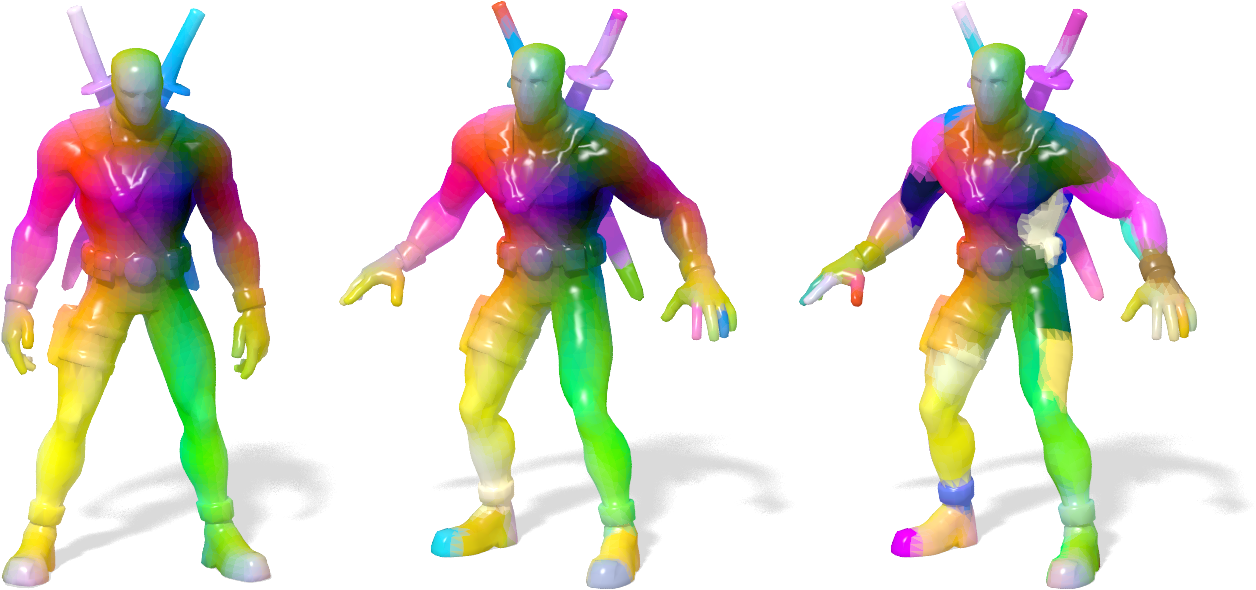}
    \includegraphics[width=0.26\linewidth]{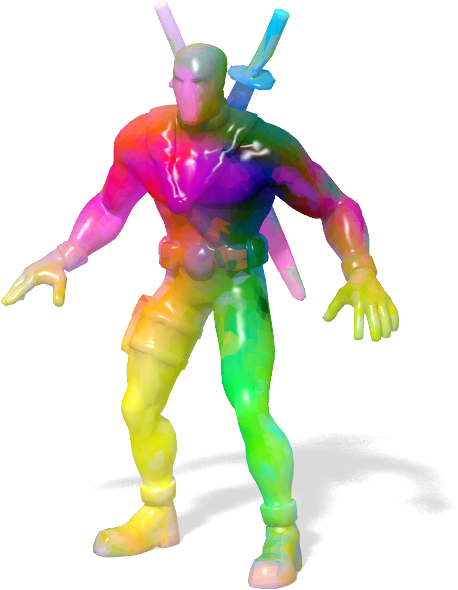}
    \includegraphics[width=0.71\linewidth]{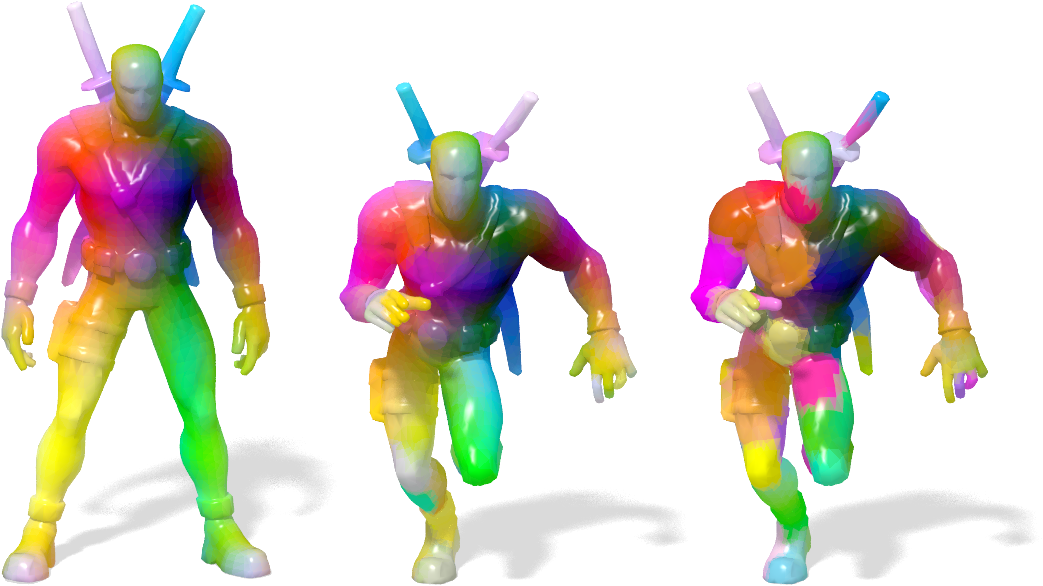}
    \includegraphics[width=0.24\linewidth]{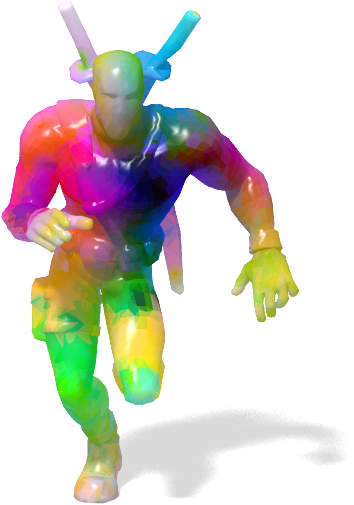}
    \includegraphics[width=0.72\linewidth]{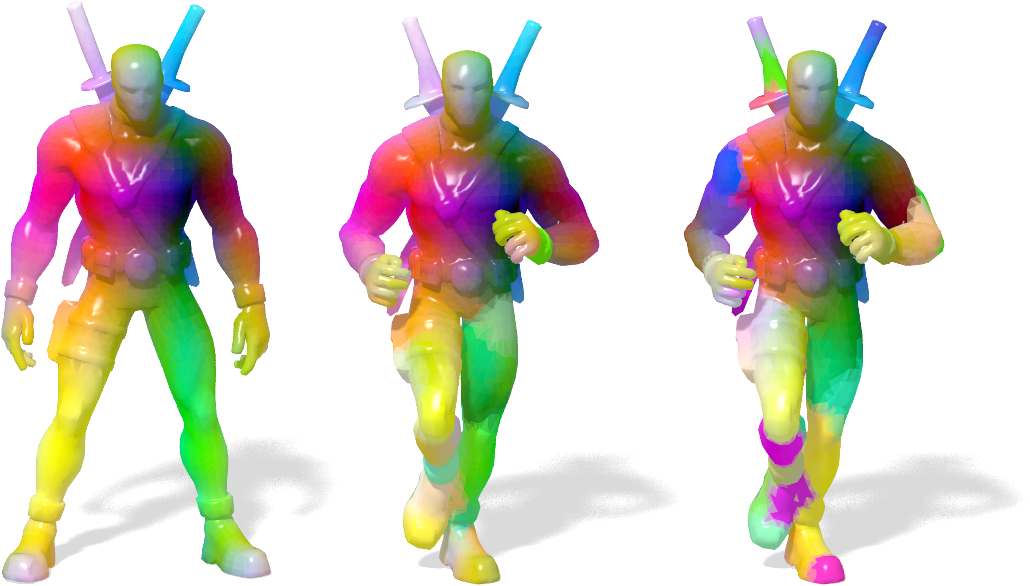}
    \includegraphics[width=0.24\linewidth]{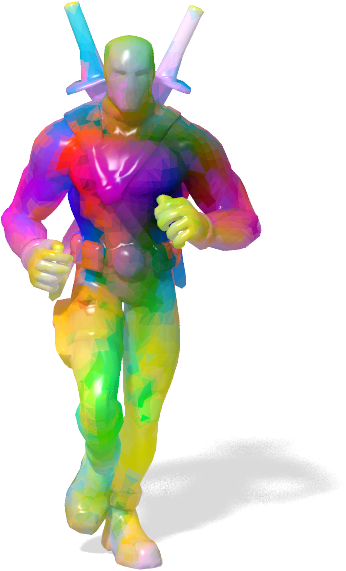}
    \includegraphics[width=0.71\linewidth]{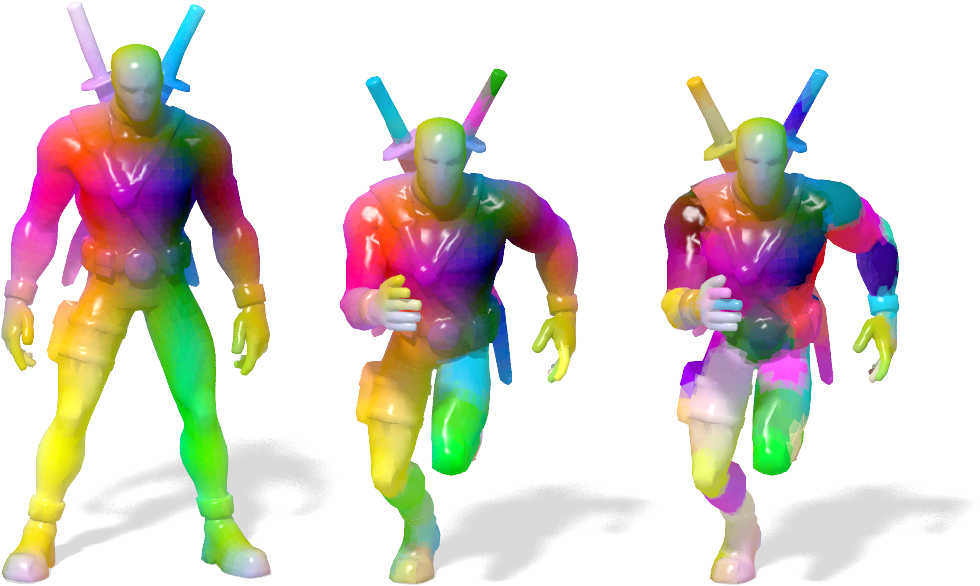}
    \includegraphics[width=0.24\linewidth]{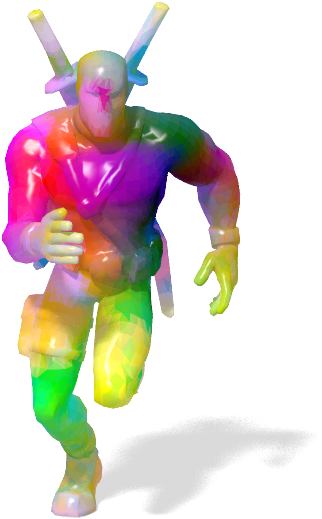}
    \caption{From left to right: Reference model; Correspondences calculated using our unsupervised network, trained on just $3$ poses of Deadpool; Correspondences calculated using a supervised network, trained on FAUST synthetic human dataset ($80$ shapes); Correspondences calculated using the purely axiomatic method of functional maps with SHOT descriptors. Note that only the axiomatic results are refined using PMF, while for the former we show the raw network predictions. Corresponding points are assigned the same color.}
    \label{fig:deadpool}
\end{figure}

\begin{figure*}[tb]
\begin{center}
    \begin{overpic}
    [trim=0cm -0.4cm 0cm 0cm,clip,width=1.0\linewidth]{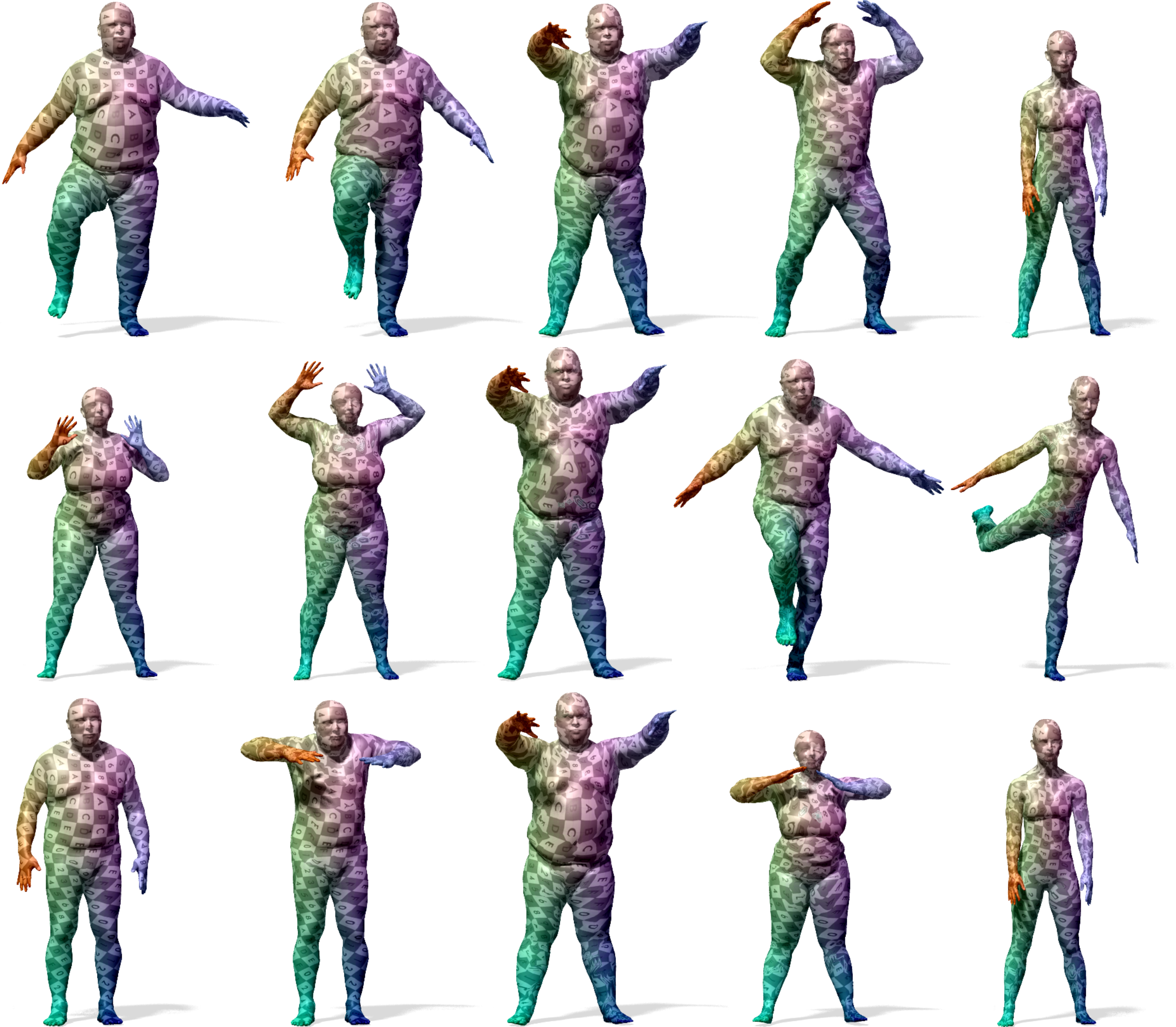}
    \end{overpic}
\caption{\label{fig:dfaust_pmf}Generalization of our network trained on synthetic Faust dataset to Dynamic FAUST~\cite{dfaust:CVPR:2017}, illustrated by texture transfer according to the estimated map. In order to convert our method's raw outputs to a bijection, results are refined using PMF~\cite{vestner2017product}. In each row, the first column shows the reference shape to which the remaining shapes are matched.}
\end{center}
\end{figure*}

\section{Training with scarce data}
\label{app:scarce}
As discussed in Section {\color{red}5.1}, having an unsupervised learning method bridges the gap between axiomatic solvers and supervised learning methods. The latter can excel on particular data but often suffer from limited generalization capabilities, while the former offer a general purpose tool for solving matches between unseen pairs but suffer from computational inefficiency. Our method can do both, while demonstrating improved capabilities in both regimes. The TOSCA experiment in Section {\color{red}5.4} (Figure {\color{red}6}) shows that our network trained on human scans generalizes well to non-human shapes. The single-pair experiment in Section {\color{red}5.1} (Figure {\color{red}1}) shows that our method can efficiently solve for a single unseen pair of shapes. In addition, we have emphasized its usefulness in fast inference given a scarce unlabeled train set. In what follows we present additional evidence that did not fit into the main paper due to page limitation.  

\vspace{1ex}\noindent\textbf{Learning to match a single pair.}
The extreme, one-pair settings described in Section {\color{red}5.1} of the main manuscript, compared our method and a supervised learning-based one. Here Figure~\ref{fig:hero} provides an additional comparison with an axiomatic method: Functional Maps~\cite{ovsjanikov12} with SHOT descriptors~\cite{SHOT}, refined with PMF~\cite{vestnerefficient}. Interestingly, this computationally intensive method is still inferior to ours. 

\vspace{1ex}\noindent\textbf{Fast inference.}
As discussed above, when fast inference is required on newly encountered unlabeled data, \comment{and only a few input shapes are given}axiomatic methods are no longer an option. Also, one cannot afford full retraining and therefore has two options: either using a pre-trained network on a labelled similar data using supervised learning, or use the unsupervised network to train quickly on few examples. We demonstrate this using an artistic model of {\em Deadpool}, a super-hero comics character, provided in a variety of poses sampled from animations.

To convert the artistic mesh to a manifold we used \cite{huang2018robust}. The models were remeshed to a $7$K resolution, using edge contraction \cite{garland1997surface}. We wish to stress that the artistic models do {\em not} have any ground-truth labeling, emphasizing the usefulness of an unsupervised approach.

In Figure~\ref{fig:deadpool} we compare the performance of the unsupervised network, trained with only $3$ shapes for a total of $15$ minutes ($100$ iterations); and the supervised network trained on FAUST synthetic human dataset ($80$ shapes) for $8$ hours ($3$K iterations). Visualized are the test examples (\ie, pairs of shapes unavailable to the network at training time). While both methods demonstrate equivalent inference time of less than one second, the performance gap is significant showing a clear advantage for our method.

\section{MPI Dynamic FAUST additional visualization}
\label{app:DFAUST}
Section 5.4 discusses the generalization of our network, trained on the FAUST synthetic human dataset, on the recent Dynamic FAUST dataset~\cite{dfaust:CVPR:2017}. As demonstrated in Figure $8$, when tested on $256$ test pairs comprising $4$ different subjects at $4$ different poses, our method showed extremely good results. To save space, we only included few visualizations. In Figure \ref{fig:dfaust_pmf} we show many more results via texture transfer.  